\documentclass{article}

\usepackage[dvipsnames]{xcolor}

\usepackage{microtype}
\usepackage{graphicx}
\usepackage{subfigure}
\usepackage{booktabs} 
\usepackage{multirow}
\usepackage{algorithmicx}
\usepackage{algorithm}
\usepackage{algpseudocode}
\usepackage{mathtools, nccmath}
\usepackage[disable]{todonotes}

\usepackage{hyperref}

\usepackage[utf8]{inputenc} 
\usepackage[T1]{fontenc}    
\usepackage{url}            
\usepackage{amsfonts}       
\usepackage{nicefrac}                
\usepackage{colortbl}
\usepackage{soul}
\usepackage{wrapfig}
\definecolor{Gray}{gray}{0.9}
\sethlcolor{Gray}

\newcommand\blfootnote[1]{
    \begingroup
    \renewcommand\thefootnote{}\footnote{#1}
    \addtocounter{footnote}{-1}
    \endgroup
}

\usepackage[accepted]{mlsys2025}

\mlsystitlerunning{Efficient LLM Inference using Vector Quantization on Mobile CPUs}

\begin{document}

\twocolumn[
\mlsystitle{GPTVQ: The Blessing of Dimensionality for LLM Quantization}

\mlsyssetsymbol{equal}{*}

\begin{mlsysauthorlist}
\mlsysauthor{Mart van Baalen}{equal,qc}
\mlsysauthor{Andrey Kuzmin}{equal,qc}
\mlsysauthor{Ivan Koryakovskiy}{qc}
\mlsysauthor{Cedric Bastoul}{qc}
\mlsysauthor{Peter Couperus}{qc}
\mlsysauthor{Eric Mahurin}{qc}
\mlsysauthor{Tijmen Blankevoort}{qc}
\mlsysauthor{Markus Nagel}{qc}
\mlsysauthor{Paul Whatmough}{qc}
\end{mlsysauthorlist}

\mlsysaffiliation{qc}{Qualcomm AI Research\thanks{\scriptsize Qualcomm AI Research is an initiative of Qualcomm Technologies, Inc.}}
\mlsyscorrespondingauthor{Mart van Baalen}{mart@qti.qualcomm.com}

\mlsyskeywords{LLMs, vector quantization, efficient inference}

\vskip 0.3in

\begin{abstract}
Large language model (LLM) accuracy and token generation rate are fundamentally limited by both DRAM footprint and bandwidth, making them challenging to deploy on mobile devices.
To improve the size-accuracy trade-off criticial to mobile LLMs, Vector Quantization (VQ) has recently been proposed.
However, while previous VQ methods demonstrated footprint reduction, they have failed to demonstrate token rate gains over standard INT4 quantization on Nvidia GPUs, and do not even consider mobile devices.
The reason for this 
is the large codebooks, which are too slow to index at inference time.
In this work, we co-design our VQ representation, post-training quantization flow, and LLM software inference engine, to enable efficient inference on mobile devices.
We propose a novel post-training quantization algorithm, GPTVQ, that quickly and accurately compresses a wide range of LLMs, specifically resulting in small per-block LUTs, which are fast to decode using existing CPU LUT instructions.
Using a custom LLM software inference engine, we demonstrate VQ LLMs running on mobile CPU, and measure a simultaneous DRAM footprint reduction of 19\% and token rate improvement of 10\% compared to industry standard INT4, at little harm to accuracy, and outperforming llama.cpp.
Finally, for task-specific scenarios, we demonstrate that combining GPTVQ base model with the orthogonal approach of LoRA adapters results in a significant improvement of accuracy over previous adapter-based methods.

\end{abstract}
]

\section{Introduction}

\blfootnote{
Snapdragon branded products are products of Qualcomm Technologies, Inc. and/or its subsidiaries.\\
\textsuperscript{*} Equal contribution.\\
\textsuperscript{1} Qualcomm AI Research (Qualcomm AI Research is an initiative of Qualcomm Technologies, Inc).
}

Large language models (LLMs) enable unprecedented improvements in usability on mobile devices, providing general AI assistance across a broad swathe of natural language processing use cases.
They also form the backbone for multi-modal models that recognize and interpret images~\cite{liu2023llava,lin2023vila}, transcribe and analyze audio~\cite{zhang2023speechgpt}, and even process video content~\cite{lin2023video,damonlpsg2024videollama2}, making them a central and indispensable tool in modern computing.

However, the sheer size of LLMs makes them challenging to deploy on mobile devices for two reasons, both pertaining to DRAM main memory constraints.
Firstly, the required model footprint is prohibitive in mobile devices, limiting achievable accuracy.
Typical mobile phones have around 8GB of total DRAM memory
\cite{wiki:iphone}, with the OS and active apps easily occupying more than half of this, typically leaving less than 4GB for an LLM.
Secondly, DRAM bandwidth limits achievable token generation rates,
since the autoregressive nature of LLMs requires loading every single weight once for each generated token.
This is particularly severe in the common case of moderate
context length of up to 8K--16K tokens~\cite{kim2023squeezellm,hooper2024kvquant}.
\textit{Therefore, reducing the stored model footprint is critical to relaxing both of these impediments.}

To directly address the compute-bound nature of LLM token generation, we explore how to trade a small increase in (surplus) compute for a commensurate decrease in valuable weight footprint and bandwidth.
We propose to do this using Vector Quantization (VQ)
~\cite{stock2019and,tseng2024quip,egiazarian2024aqlm,malinovskii2024pvtuning}, which is a SOTA approach that uses a non-uniform number system in multiple dimensions to aggressively reduce LLM footprint.
Since we cannot compute on the VQ encoded data directly, we must first decode to a \textit{native} data type.
Hence, we use VQ only as a \textit{storage} data type.

\begin{figure*}[t]
    \centering
    \includegraphics[width=0.7\linewidth]{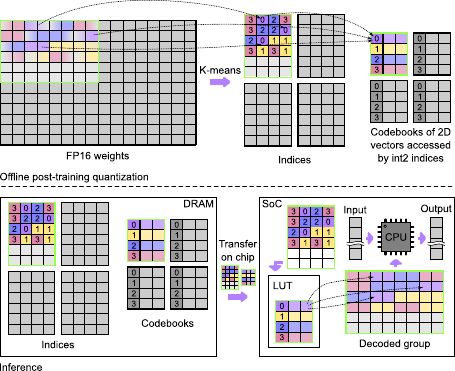}
    \caption{
    The proposed hardware-friendly representation and GPTVQ method.
    \textbf{Top:} During quantization, the FP16 weights are split into groups with their own small codebook.
    \textbf{Bottom:} During inference, the codebooks and indices are moved from DRAM to SoC independently from each other.
    The codebook is implemented as a lookup table (LUT) available on modern mobile CPUs.}
    \label{fig:vq_representation}
\end{figure*}

\todo{PW: I need to mention importance/prevalence of CPU for mobile somewhere}
VQ provides an improvement in model size vs accuracy, which will ideally improve all of: footprint, bandwidth and token rate. 
However, although the token rate improvement is expected since the workload is heavily bandwidth bound, it is not guaranteed, due to the overhead for decoding VQ back to a native type before use.
If carefully designed, this overhead
is more than offset by the reduction in memory bandwidth, enabling either a larger number of parameters (higher accuracy) at the same DRAM footprint and token rate, or a higher token rate and smaller footprint for the same number of parameters.
However, previous research on VQ targets cloud GPU platforms only, and uses very large codebooks, which are not efficient for implementation on mobile CPUs.
\todo{PW: do we have any experiments to show decode latency vs LUT size?}

In fact, we find that VQ decoding is most efficiently implemented on mobile devices using existing hardware \textit{lookup table instructions}, present on mobile CPUs and also on many NPUs and GPUs.
These instructions typically map a 5- or 6-bit index to an 8-bit value.
Critically, this means that very large lookup tables or high VQ dimensions, such as those used by AQLM~\citep{egiazarian2024aqlm}, require many calls to the lookup table instruction, leading to significant decoding latency.
In this work, we demonstrate the potential of VQ footprint compression running on mobile CPU, by co-designing the VQ compression algorithm with the software implementation.

\todo{need to be careful here, can we show this concretely?}
Figure~\ref{fig:vq_representation} demonstrates our approach.
During post-training quantization, we divide the FP16 weights into groups.
For each group, a table of indices and a corresponding codebook is derived.
During inference, the indices and codebooks are stored in DRAM, before being transferred to the CPU cache on the SoC, where they are decoded to a native data type and used in matrix multiplication. 
To minimize the quantization error due to our VQ representation, we introduce a novel post-training quantization method, \textit{GPTVQ}.
The quantization of each weight matrix is implemented as a single pass from left to right, which is highly efficient. The quantization error is compensated by updating the remaining unquantized weights.
Finally, we implement a custom LLM inference software engine for mobile CPU, to demonstrate that our end-to-end VQ method can reduce footprint and increase token rate, without compromising accuracy.

The contributions of this work are summarized as follows:
\begin{itemize}
    \item 
    We describe an optimized VQ representation that is not only footprint efficient, but also fast to decode.
    This is achieved by co-designing the VQ parameters, including the bitwidths, number of dimensions and LUT size, with the mobile CPU software implementation, to efficiently leverage existing hardware ISA extensions for fast LUT decoding.
    
    \item We implement and benchmark a full LLM inference stack supporting VQ decompression on mobile CPU.
    Results demonstrate that VQ reduces DRAM footprint by 19\%
    while increasing the token rate by 10\%, 
    compared to a 4-bit integer baseline.
    
    \item We propose a fast and accurate algorithm for post-training VQ compression (GPTVQ), 
    which achieves favorable size vs accuracy trade-offs on a wide range of LLMs, while having a practical offline run time of only 3 to 11 hours on a 70B parameter model.
    
    \item We also show that our VQ approach is complementary to the popular use of adapters with LLMs, which gives an additional opportunity to recover accuracy loss from aggressive quantization on mobile devices.

\end{itemize}

\section{Background and Motivation}

In this section we motivate the use of VQ as a storage data type.
We will first establish notation for VQ and explain why VQ provides better representational accuracy than traditional uniform quantization.
Then, we will discuss existing VQ methods and their drawbacks.
Lastly, we discuss requirements of on-device implementation of VQ storage type decoding to a native compute data type, and how existing approaches aimed at cloud GPU environments fall short for mobile application.

\subsection{Uniform, non-uniform, and vector quantization}\label{sec:quant_background}

A symmetric uniform quantizer approximates an original floating point vector $\textbf{x} \in \mathbb R^{D}$ as $\textbf{x}\approx~s \textbf{x}_{int}$, where  each element in $\textbf{x}_{int}$ is a $b$-bit integer value and $s$ is a higher precision quantization scale, shared across the components of $\textbf{x}$. 

A more flexible quantization approach is non-uniform quantization, in which floating point numbers are discretized to arbitrary scalar centroids stored in a codebook $C:C=\{c_1,c_2,\dots,c_k\}$. 
Each high precision value in $\textbf{x}$ is then represented by the index $j$ of a centroid $c_j$. 
Each index is stored in $\lceil \log_2 k \rceil$ bits.
Even more flexible quantizer can be constructed using a higher-dimensionality for the centroids of $C$. 
In this case, each centroid in $C$ encodes $d$ values, e.g., pairs of values if $d=2$, and each group of $d$ values in $\textbf{x}$ is represented by a single index into $C_d$, where $C_d$ denotes a codebook with elements of dimensionality $d$~\cite{gersho2012vector}.

\begin{figure}[t]
    \centering
    \begin{subfigure}
        \centering
        \includegraphics[width=0.45\textwidth]{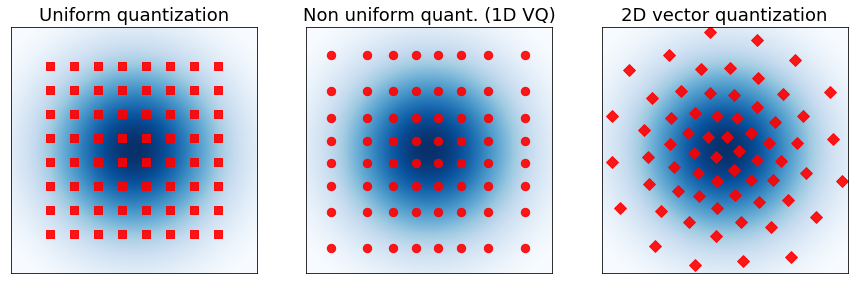}
    \end{subfigure}
    \begin{small}
        \begin{tabular}{c}
        \includegraphics[width=0.4\textwidth]{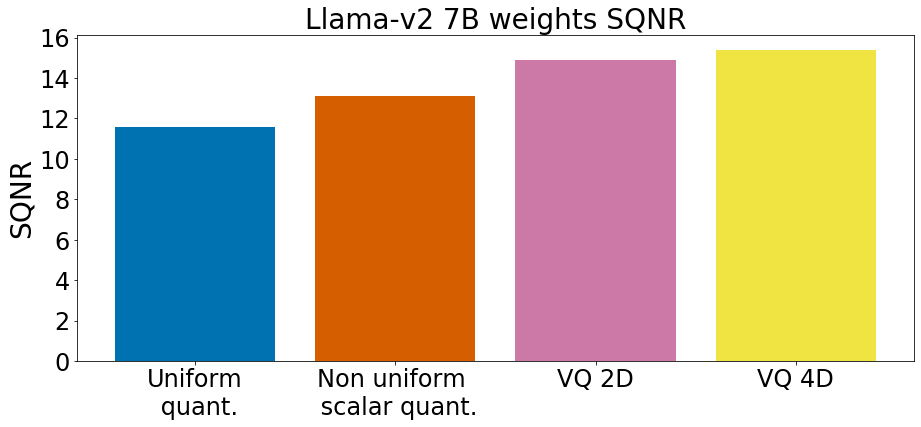} 
        \end{tabular}
    \end{small}
    \caption{
    \textbf{Top:} Illustration on how vector quantization can fit better 2D normal data, compared to uniform and non-uniform grids. 
     \textbf{Bottom:} SQNR increases with quantization dimensionality on Llama-v2 7B weights, due to additional flexibility in the quantization grid.}
    \label{fig:quant_sqnr}
\end{figure}

Increasing the dimensionality of the codebook via VQ, increases the flexibility of the quantization grid. Figure~\ref{fig:quant_sqnr}~\textit{(top)} gives a visual representation of this.
In this example, where we quantize each value in the original to a 3-bit representation, i.e., 6 bits for each pair of values, we can see that the number of points stays the same, i.e., $2^6 = 64$, but the distribution of the centroids with VQ can more closely match the underlying distribution, increasing the accuracy of the representation. 

The representational accuracy increases the more the dimensionality of the codebook increases.
Figure~\ref{fig:quant_sqnr}~\textit{(bottom)} shows signal-to-quantization-noise ratio (SQNR) as the accuracy measure.
SQNR is defined in the log scale as: $\text{SQNR}_{dB}=10\log_{10}\left(\mathbb E\left[ W^2\right]/  \mathbb E \left[ (W-Q(W))^2 \right]\right)$,
where $Q(\cdot)$ is the quantization function.
We can see the improvement in representational accuracy of higher $d$ for Llama-v2 7B weights. 
Note that, as $d$ grows, so does the codebook size. For fair comparison, we ensure the codebook overhead is always equal to 0.25 bit per weight for each quantization method, i.e., improved SQNR is not caused trivially by using more bits in total for our representations.

\subsection{Prior work on vector quantization}\label{sec:vq_challenges}

\todo{MN: IMO this is to detailed, should that not be in the appendix? For the story only the part on the representation seems relevant.}
Due to its ability to represent data more flexibly at very low bit-widths, vector quantization has received increased attention recently.
The AQLM method introduced in \cite{egiazarian2024aqlm} employs vector quantization to compress LLMs down to approximately 2 bits per weight, with significantly improved accuracy compared to uniformly quantized models at the same bit width.
Their method consists of 3 stages. 1) A codebook initialization step, where a weight tensor is reshaped into a matrix with $d$ columns and the codebook is initialized using k-Means; 2) 100 iterations of an EM-style phase, in which one epoch of a gradient-based codebook fine-tuning is followed by a beam search that updates codebook indices to minimize layer-wise reconstruction error; and 3) a full block fine-tuning step in which the entries of the codebooks for all layers in a decoder block are fine-tuned to minimize the output error of the block.
AQLM uses vector dimensions $d=8$.
At 2 bit per dimension, this means that each codebook contains $2^{16}$ 8-bit vectors.

While AQLM shows good quantization accuracy, the EM-style phase and the following block fine-tuning are expensive to run.
In our experiments on a single H100, Llama v2-7B quantization takes approximately 35 hours.

More recently, PV-tuning \cite{malinovskii2024pvtuning} introduces an end-to-end training method for non-uniform and VQ quantization.
Similar to uniform quantization, gradients cannot be backpropagated through centroid indices.
In uniform quantization, a method commonly used to circumvent this issue is the \emph{straight-through estimator} (STE). With STE, during the backward pass, the gradient of the loss with respect to the quantized weights is passed 'straight-through' to the unquantized (shadow) weights without modification.
The PV-tuning authors show that this approach does not work for VQ, as updating all weights simultaneously leads to too-large updates.
Instead, to learn centroid indices, the authors introduce a trust ratio that restricts weight updates to a small subset.
Using this method the authors show improved model accuracy compared to models quantized using AQLM.

\subsection{VQ model deployment and limitations of existing methods}\label{sec:vq_deployment_challenges}

To avoid introducing extra latency in the decoding step from storage data type to compute data type, 
 decoding needs to be extremely efficient and faster than the DRAM bandwidth.
The most efficient way to decode VQ weights is by using the \emph{lookup table instruction} (LUT) which is present in all modern mobile CPUs.
The LUT instruction maps a 6-bit index to an 8-bit value.
This means that, for 2D VQ, 2 LUT instructions must be called, one for each dimension.
The 6-bit index implies that VQ codebooks should contain at most 64 entries.
For this reason the setting used by recent VQ methods such as AQLM is not conducive to good on-device performance: Using, e.g., 16-bit indices (the 2.29 bpv in \cite{egiazarian2024aqlm}) precludes the use of the efficient LUT instruction, and instead requires the use of the less performant SVE gather instruction.

Armed with this knowledge, we design a VQ representation using fewer bits per index and lower dimensionality, and show that this can achieve model accuracy competitive with traditional INT4 quantization and other VQ approaches. Furthermore, we show that this setting can be implemented efficiently for inference on mobile CPU.
\section{On-device implementation}\label{sec:method_on_device_impl}
In this section we describe our on-device implementation.
\todo{MN: we need a better title here. Or we make two sections of it?}

\subsection{Mobile-friendly quantized tensor representation}

\todo{This needs proof reading to check for repetition/overlap}
\todo{MN: I think here we need more details, e.g. mention we use 6 (or even 4) bit indices, just 2D, all to reduce the number of LUT calls. Also the tiling we might wanna mention a bit more and detailed.}
The most common approach to LLM quantization on mobile CPU platforms is to use 4-bit integer for each element in a tensor with a scale factor shared among a group of elements.
This approach is widely adopted, e.g., by the open source Llama.cpp\footnote{\url{https://github.com/ggerganov/llama.cpp}} project.
Our VQ implementation follows a similar scheme, with tensors split into groups of elements, each with a scale factor.  
In addition, we store for each block a lookup table that is used to decode VQ elements.
Hence, each group of weight elements is stored as a tuple of 1) the VQ encoded element indices, 2) the associated LUT for decoding, and 3) a quantization scale factor.
While the GPTVQ algorithm presented in the next section can be applied to any dimensionality or index bitwidth, for our practical implementation we choose a fixed configuration of 2D VQ, with 6-bit indices, i.e., 3 bits per weight. 
This allows our LLM inference software implementation to leverage the 6-bit to 8-bit LUT instruction.

\subsection{LLM Inference Software Implementation} 
The inference software we use to benchmark our VQ approach is an in-house implementation of a highly parameterizable transformer architecture that supports major large language models. 
It is written in C with vector intrinsics for accelerating matrix multiply and similar kernels on mobile CPU using SIMD extensions and the like.
Furthermore, it leverages the structural properties of transformers for efficient coarse-grain parallelization and high-level polyhedral compiler capabilities for fine-grain vectorization. 

To support VQ in the inference engine, the 6-bit indices are packed tightly and stored in memory along with the lookup tables and quantization scales, organized to enable efficient vectorization.
During inference, each block is decoded as follow: The tuple of block data is loaded from DRAM onto the SoC, and into the CPU cache, c.f. Figure~\ref{fig:vq_representation}.
Here, the VQ decode kernel uses the native mobile CPU LUT instructions to efficiently perform lookups quickly, converting 6-bit index to signed 8-bit integer data.
These integer values are then used in the downstream matrix-vector multiplications.

\section{The GPTVQ algorithm}\label{sec:method}
In this section, we introduce our GPTVQ algorithm, a novel method for efficient and accurate post-training vector-quantization of LLMs, which extends the GPTQ \cite{frantar2022gptq} algorithm to VQ.
Appendices~\ref{sec:codebook_update},~\ref{sec:em_init}~,\ref{app:codebook_compression},~and~\ref{sec:blockwise_data_normalization} present extensions to GPTVQ, including Codebook SVD, Blockwise Data Normalization, an extended EM initialization algorithm, and a codebook update procedure.

Neural network quantization reduces model size as well as compute and energy requirements, but introduces quantization noise. 
A large body of literature exists with methods to alleviate the effects of quantization noise on model accuracy, see \cite{nagel2021whitepaper, gholami2022survey} for recent surveys.
Post-training quantization (PTQ) approaches aim to mitigate the adverse effects of quantization noise on pre-trained networks, without having to resort to costly quantization-aware training (QAT).
A popular and effective approach in PTQ, introduced by AdaRound \cite{nagel2020adaround}, is to modify weights to minimize a layer's output error as an approximation to the full network's loss:
\begin{equation}\label{eq:layerwise_loss}
\mathbb{E}\left[\mathcal{L}(\mathbf{\theta}+\mathbf{\epsilon})-\mathcal{L}(\mathbf{\theta})\right]\approx\sum_{\ell}||\mathbf{W}^{\ell}\mathbf{X}^{\ell}-\widehat{\mathbf{W}}^{\ell}\mathbf{X}^{\ell}||^2_F, 
\end{equation}
where $\mathbf{W}^{\ell}$ is the weight for layer $\ell$, $\widehat{\mathbf{W}}^{\ell}=\mathbf{W}^{\ell}+\mathbf{\epsilon}^{\ell}$ is the (quantized) approximation to this weight tensor, and $\mathbf{X}^{\ell}$ of shape $R \times N$ denotes the input data for layer $\ell$ from a calibration dataset, with $N$ individual data points of dimensionality $R$ along its columns.

\paragraph{Preliminary: GPTQ}

\begin{algorithm}[t]
	\caption{GPTVQ algorithm: Quantize a weight tensor $\mathbf{W}\in\mathbb{R}^{r\times c}$ given the inverse Hessian $\mathbf{H^{-1}}$, the block size $B$, VQ dimensionality $d$, the number of centroids $k$, and the group size $l$. To simplify the notation, we assume one group per column. } 
 
\begin{algorithmic}[1]
            \State $N_b \leftarrow\frac{c}{B}$\Comment{the number of blocks}
            \State $m \leftarrow\frac{l}{r}$\Comment{the number of columns in a group}
            \State $\mathbf{Q}\leftarrow\mathbf{0}_{r,c}$
            \State $\mathbf{E}\leftarrow\mathbf{0}_{r,c}$
            \State $N_g \leftarrow\frac{rc}{l}$\Comment{the number of groups/codebooks}
            \State $\mathbf{C}_{i}\leftarrow \mathbf{0}_{d,k}, i=1,\ldots,N_g$
            \State $\mathbf{H}^{-1}\leftarrow \text{Cholesky}(\mathbf{H}^{-1})^T$
		\For {$i=0,B,2B,\ldots, N_b B$}
                \If{i \% m = 0}
                    \State $g \leftarrow \frac{i}{m} $ \Comment{the group index}
                    \State $\mathbf{C}_g\leftarrow\text{init\_codebook}\left[ \mathbf{W}_{:,i:i+m-1} \right]$
                \EndIf
                \State $\mathbf{Q}_{:,i:i+m-1}$ $\leftarrow$ \Call{QuantGroup}{$\mathbf{W}_{:,i:i+m-1}$}
                \State $\mathbf{W}_{:, i+B:}\leftarrow  \mathbf{W}_{:, i+B:} -\mathbf{E} \cdot [\mathbf{H}^{-1}]_{i:i+B, i+B:}$
		\EndFor
\end{algorithmic} 
\label{gptvq_algo_flow}
\end{algorithm}
\begin{algorithm}
	\caption{QuantGroup: VQ quantization for a group of weights $\mathbf{W}\in\mathbb{R}^{r\times m}$ given the inverse Hessian $\mathbf{H^{-1}}$, the block size $B$, VQ dimensionality $d$} 
\begin{algorithmic}[1]
        \Function {QuantGroup}{$\mathbf{W}$}
            \For {$j=0,d, 2d, \ldots,l$}
                \State $P={{j,\ldots,j+d-1}}$
	           \State $\mathbf{Q}_{:,P}\leftarrow  \text{VQ\_quant}\left[\mathbf{W}_{:,P}, \mathbf{C}_g\right]$ \label{lst:line:quant_op}
                \State $\mathbf{E}_{:, P}\leftarrow \left( \mathbf{W}_{:,P}-\mathbf{Q}_{:,P} \right)[\mathbf{H}^{-1}]_{P}$ \label{lst:line:ref_line}
                \State $\mathbf{U} \leftarrow \sum_{p=0}^{d-1}\mathbf{E}_{:,j+p} [\mathbf{H}^{-1}]_{p,j+d-1:B}$
                \State $\mathbf{W}_{:, j+d-1:B}\leftarrow  \mathbf{W}_{:, j+d-1:B} - \mathbf{U}$
                \label{lst:line:ref_accum_update}
            \EndFor
        \EndFunction
	\end{algorithmic} 
\label{gptvq_algo_single_group}
\end{algorithm}
GPTQ follows Optimal Brain Quantization (OBQ; \cite{frantar2022optimal}), which uses the Hessian of Equation~\ref{eq:layerwise_loss}. \todo{MN: which Hessian in eq1? It is in the adaround paper but not in the equation...}
This Hessian can be efficiently computed as $\textbf{H}^{(\ell)}=\textbf{X}^{(\ell)}\textbf{X}^{(\ell)T}$.
Like OBQ, GPTQ aims to minimize the Hessian-weighted error introduced by quantizing weights in $\textbf{W}^{(\ell)}$:
\begin{align}\label{eq:gptq_error}
E = \sum_q |E_q|_2^2; &  & E_q=\frac{(\mathbf{W}_{:,q}-\text{quant(}\mathbf{W}_{:,q}))^2}{\left[\textbf{H}^{-1} \right]_{qq}}.
\end{align}
GPTQ extends OBQ in the following ways.
First, GPTQ exploits the fact that $\textbf{H}^{(\ell)}$ is shared over all rows of $\textbf{W}^{(\ell)}$ by quantizing all weights in a column in parallel, from left to right.
This obviates the need for independent Hessian updates for different rows.
After quantizing a column $q$, all remaining (unquantized) columns $q'> q$ are modified with a Hessian-based update rule $\mathbf{\delta}$ that absorbs the error introduced by quantizing column $q$ on the layer's output:
\begin{equation}\label{eq:update_rule}
\mathbf{\delta}=-\frac{\mathbf{W}_{:,q}-\text{quant(}\mathbf{W}_{:,q})}{\left[\textbf{H}^{-1} \right]_{qq}}\textbf{H}_{:,(q+1):}
\end{equation}
To reduce data transfer, GPTQ applies the update of Equation~\ref{eq:update_rule} only to a small block of $B$ columns in which column $q$ resides.
Note that such blocks enable an efficient weight update implementation.
To update the columns outside of block $B$, the error $E_q$ in Equation~\ref{eq:gptq_error} is accumulated while the columns in block $B$ are processed, and are applied in one go to all columns outside of block $B$ after all columns in block $B$ are processed.
Lastly, GPTQ uses a Cholesky decomposition for updating the inverse Hessian $\textbf{H}^{-1}$, which introduces a more numerically stable alternative to the inverse Hessian row and column removal operations of OBQ.

\paragraph{Codebook Initialization}
Unlike traditional GPTQ, VQ requires a codebook for each group of weights. 
To initialize the codebook for a group of weights, we propose the following variant of the EM algorithm. Given the set of $d$-dimensional vectors $\mathbf{x}^{(i)}$, our goal is to find $k$ centroid vectors $\mathbf{c}^{(m)}$ and the corresponding sets of indices $I_{m}$ pointing at the centroid $m$. The objective is the following sum of weighted distance functions:

\begin{equation}
\label{eq:kmeans_objective}
    \min_{\mathbf{I},\mathbf{c}^{(0),\dots,(k)}} \sum_{m=0}^{k} \sum_{i\in I_{m}} \left(\textbf{x}^{(i)} - \textbf{c}^{(m)}\right)^T \textbf{D}^{(i)}\left(\textbf{x}^{(i)} - \textbf{c}^{(m)}\right),
\end{equation}
where $\mathbf{D}=\text{diag}\left(1/[\mathbf{H}^{-1}]_{11}, \dots, 1/[\mathbf{H}^{-1}]_{cc}\right)$ and $\mathbf{D}^{(i)}$ is a $d\times d$ subset of $\mathbf{D}$ corresponding to the data point $\mathbf{x}^{i}$. E.g. for 2D vector quantization, these matrices are share among pairs of columns. For the case of $\textbf{D}^{(i)}$ equal to identity, the clustering method is equivalent to K-means. The objective can be minimized using E- and M-steps as follows. 

\textbf{E-step}: keeping the values of the centroids fixed, find centroid assignment $\textbf{c}^{(i)}$ for each unquantized $d$-dimensional vector $\textbf{x}^{(i)}$ that minimizes the objective~(\ref{eq:kmeans_objective}). Using this distance function assigns optimal centroids based on the data-aware loss.

\begin{equation}
\label{eq:hessian_e_step}
    \textbf{c}^{(i)}=\arg\min_{m}  \left(\textbf{x}^{(i)} - \textbf{c}^{(m)}\right)^T \textbf{D}^{(i)}\left(\textbf{x}^{(i)} - \textbf{c}^{(m)}\right),
\end{equation}

\textbf{M-step}: keeping the assignments found in the E-step fixed, find the centroid value $\textbf{c}^{(m)}$ that minimizes
\begin{equation}
\label{eq:hessian_m_step}
    \textbf{c}^{(m)} = \arg\min_{\textbf{c}} \sum_{i \in I_{m}}\left(\textbf{x}^{(i)} - \textbf{c}\right)^T\textbf{D}^{(i)}\left(\textbf{x}^{(i)} - \textbf{c}\right).
\end{equation}
This objective is a quadratic form w.r.t $\mathbf{c}^{(m)}$. The optimal value is computed in a closed form as $\textbf{c}^{(m)}=\left(\sum_{i \in I_{m}}\textbf{D}^{(i)}\right)^{+}\left(\sum_{i \in I_{m}}\textbf{D}^{(i)}\textbf{x}^{(i)}\right)$, where $(\cdot)^+$ is a Moore–Penrose pseudoinverse.
During the vector quantization operation on line 4 in Algorithm~\ref{gptvq_algo_single_group}, we use the assignment step defined in Equation~\ref{eq:hessian_e_step} as well.
Practically, we find no performance difference between using the inverse Hessian diagonal, or the full $d$-dim inverse sub-Hessian.

\paragraph{GPTVQ}
To efficiently and accurately quantize pre-trained floating point LLMs to our VQ storage format, we generalize the GPTQ framework to non-uniform and vector quantization.

Following the GPTQ framework we perform quantization of the weight tensor in a greedy manner starting from the first column. The details of the method are given in Algorithm~\ref{gptvq_algo_flow}.  For a VQ dimensionality $d$, we quantize $d$ columns at a time. Although quantizing $d$ elements of a single column instead of $d$ columns would absorb the quantization error more effectively, we choose to quantize $d$ columns at a time as this layout allows for a much more efficient on-device implementation.

When quantizing a $d$-dimensional vector, we find the centroid in the codebook for the current block that minimizes the objective in Equation~\ref{eq:kmeans_objective}.
After quantizing $d$ columns, we update the remaining weights using the update rule~\eqref{eq:update_rule}. We accumulate the update along $d$ coordinates and apply it to the remaining weights as a single operation. To allow for a hardware-friendly representation, we use several codebooks per layer, each assigned to a \textit{group} of weights (see Algorithm~\ref{gptvq_algo_flow}).
We use group sizes of at most 256 columns, to ensure codebook initialization can capture the previous updates of~\eqref{eq:update_rule}. For example, a group of 2,048 weights is 8 rows by 256 columns.

\paragraph{Total bits per value} 
As a measure of total model size, we compute \emph{bits per value} (bpv), given by $\log_2(k)/d+kdb_c/l$ 
, where $k$ is the number of centroids, $d$ is the $VQ$ dimensionality, $b_c$ is the codebook bit-width, and $l$ is the group size, i.e., the number of weights sharing a codebook. 
We choose values for $k$ s.t. $\log_2(k)$ is an integer.

\section{Experiments and results}
\label{sec:experiments}

In this section we compare on-device token generation rate of our VQ method to uniformly quantized models. Furthermore, we evaluate GPTVQ and compare the performance of vector quantization in 1, 2 and 4 dimensions against uniform quantization baseline methods.

\subsection{On-device VQ inference evaluation and comparison}
To investigate the effect of VQ quantized models on model DRAM footprint and token rate, 
we run our optimized VQ decoding kernel with our 
in-house implementation of a highly parameterizable transformer architecture, as described in Section~\ref{sec:method}.
We benchmark the latency and DRAM footprint on a Snapdragon\textregistered{} X Elite platform, and further compare our implementation to the open source llama.cpp implementation baseline on the same platform.
The mobile platform we used runs Windows, 
and we used Clang 18.1 with Polly enabled.

We measure end-to-end inference token rate for a Llama3-8B model using three different quantization scenarios: 1) llama.cpp Q4\_0 INT4 quantization; 2) our implementation with INT4 g128; and 3) Our implementation with 2D VQ, at 3.125 effective bits per weight. 
The latter uses 3 bits per dimension, i.e. 6-bit indices, and a group size of 8,192.

\begin{table*}
\centering
\vspace{-.25cm}
\normalsize
\setlength{\tabcolsep}{5pt}
\caption{\textbf{
LLM token generation on mobile device for Llama-v3-8B.} The top row shows model footprint and token rate numbers for a model deployed using llama.cpp, the two bottom rows show deployment using our implementation.
}\label{tab:latency2}
\vspace{0.2cm}
\begin{tabular}{cccccc}
    \toprule
    Bits per value & Format & Block size & Engine & Footprint [GB] $\downarrow$ &  Throughput [tok/s] $\uparrow$ \\ 
    \midrule
    4.5 & INT4 & 32 & llama.cpp & 4.64 & $17.95^{\pm1.01}$ \\
    \midrule 
    4.125 & INT4 & 128 &  Ours & 4.33 & $23.81^{\pm0.27}$ \\
    3.125 & VQ 2D & 8,192 & Ours & \textbf{3.52} (\textcolor{Green}{-19\%}) & $\mathbf{26.15^{\pm0.31}}$ (\textcolor{Green}{+10\%}) \\

    \bottomrule
\end{tabular}
\end{table*}

Results of this experiment can be found in Table~\ref{tab:latency2}. Firstly, we note that our INT4 implementation baseline greatly improves token rate compared to llama.cpp.
Secondly, we note that, within our optimized implementation, VQ provides a 10\% increase in token rate compared to INT4 g128 quantization.
Lastly, we note that our VQ model has significantly lower footprint than the INT4 quantized models.

In conclusion, when deployed on a mobile CPU, our VQ setting yields significantly lower footprint and significantly higher token rates than an INT4 model, even when compared to a highly optimized INT4 implementation.

\subsection{GPTVQ evaluation}
\paragraph{Models} We use Llama-1 \cite{touvron2023llama}, Llama-2 \cite{touvron2023llama2}, and Llama-3 as well as Mistral-7B-v0.1 \cite{jiang2023mistral} and Mixtral-MoE-8x7B-v0.1 \cite{jiang2024mixtral}. 
Additionally, we run a single ablation on BLOOM-560M \cite{workshop2022bloom}. 

\paragraph{Datasets} Following \citet{shao2023omniquant}, we use 128 sequences of 2048 tokens from the WikiText2 \cite{merity2016pointer} training set as calibration data for all experiments. 
We evaluate our models on token perplexity for the WikiText2 validation set for a sequence length 2048, as well as zero-shot language tasks: PIQA \citep{bisk2020piqa}, ARC-easy/-challenge \cite{allenai:arc}, BoolQ \cite{clark2019boolq}, HellaSwag \cite{zellers2019hellaswag}, and WinoGrande \cite{ai2:winogrande}.  
For Llama3, following \cite{huang2024llama3quant}, we omit BoolQ from the zero-shot average to allow fair comparison to the zero-shot results in \cite{huang2024llama3quant}.
For all evaluation tasks except WikiText2 perplexity we use the LLM-evaluation-harness \cite{eval-harness}.

\paragraph{Baselines} We compare GPTVQ to various uniform quantization methods with different group sizes, at the same overall bits-per-value (bpv).
We include
Round-to-Nearest (RTN) and several recent state-of-the-art PTQ approaches for LLMs: GPTQ \cite{frantar2022gptq}, AWQ \cite{lin2023awq}, and OmniQuant \cite{shao2023omniquant}. 
We take AWQ and OmniQuant baseline numbers from \cite{shao2023omniquant}, all Llama3 baseline numbers from \cite{huang2024llama3quant}, and generate all other baseline numbers ourselves.
In Section~\ref{sec:other_vq} and Appendix~\ref{app:aqlm} we provide a more detailed comparison to other VQ approaches, most notably AQLM \cite{egiazarian2024aqlm}, recent work that applies VQ to LLMs in a different manner.

\paragraph{Codebook overhead}
For a given bits per index $b$ and VQ dimensionality $d$, we set group size $l$ to reach an overhead of 0.125 bits per value for all values of $b$, and additionally consider an overhead 0.25 bits per value for $b=2$. 
These are chosen to match the overhead incurred by a 16-bit quantization scale for the commonly used group size of 128 (e.g., \cite{frantar2022gptq}) and the group size of 64 used by \cite{shao2023omniquant}.

\paragraph{Comparison to scalar quantization}

\begin{table*}[t]
    \setlength{\tabcolsep}{5pt}
    \small
    \centering
    \caption{\textbf{Weight-only quantization results of Llama-v2/v3, Mistral, and Mixtral-MoE Models}. We report WikiText2 perplexity and average zero-shot accuracy; Models marked L2 denote Llama-v2, L3 denote Llama-v3, M denotes Mistral, and 8x7B denotes Mixtral-MoE 8x7B. Numbers marked in bold are SOTA or surpass it, numbers underlined are on par with or outperform at least one VQ variant. * Following \cite{huang2024llama3quant}, Llama3-8B zeroshot average omits BoolQ.}\label{tab:main_results}
    \vspace{0.2cm}
    \renewcommand{\arraystretch}{0.98}
    \setlength\tabcolsep{3pt}
    \begin{tabular}{llccccccc|ccccc}
        \toprule
        & & \multicolumn{7}{c}{WikiText2 perplexity $\downarrow$} &  \multicolumn{5}{c}{Zeroshot avg acc. $\uparrow$}\\
          & & L2-7B & L2-13B &L2-70B & L3-8B & L3-70B & M-7B & 8x7B & L2-7B & L2-13B & L3-8B$^*$ & M-7B & 8x7B \\  \midrule
        FP16 &   & 5.47 & 4.88 & 3.31 & 6.1 & 2.9  & 5.25 & 3.84 & 70.5 & 73.2 & 68.6 & 75.7 & 75.9 \\ 
        \midrule
        \multirow{7}{*}{\shortstack{2.125\\ \\W2\\g128}} 
         & RTN &  
            4e3 & 122 & 27.3 & 2e3 & 5e5  & 1e3 & 4e3 & 
            36.9 & 42.1 & 36.0 & 37.8 & 38.3\\
         & GPTQ   & 
            36.8 & 28.1 & 6.74 & 2e2 & 11.9  & 15.7 & 14.1 &
            41.4 & 46.6 & 36.2 & 41.9 & 44.5\\
         & AWQ &   
            2e5 & 1e5 & - & 2e6 & 2e6 & - & - & - & - & - & - & - \\
         & OQ  & \underline{11.1} & 8.26 & 6.55 & - & - & - & - & - & - & - & - & -  \\
         
        & \cellcolor{Gray}\textbf{Ours 1D} 
            & \cellcolor{Gray} 12.2
            & \cellcolor{Gray} \textbf{7.40}
            & \cellcolor{Gray} \textbf{5.03}
            & \cellcolor{Gray} \textbf{15.9}
            & \cellcolor{Gray} \textbf{9.37}
            & \cellcolor{Gray} \textbf{14.0}
            & \cellcolor{Gray} \textbf{8.37}
            & \cellcolor{Gray} \textbf{47.8}
            & \cellcolor{Gray} \textbf{61.8}
            & \cellcolor{Gray} \textbf{41.1}
            & \cellcolor{Gray} \textbf{42.8}
            & \cellcolor{Gray} \textbf{54.9}
            
            \\
         
         & \cellcolor{Gray}\textbf{Ours 2D} 
			& \cellcolor{Gray} \textbf{7.77}
            & \cellcolor{Gray} \textbf{6.52}
            & \cellcolor{Gray} \textbf{4.72}
            & \cellcolor{Gray} \textbf{11.3}
            & \cellcolor{Gray} \textbf{7.37}
            & \cellcolor{Gray} \textbf{7.53}
            & \cellcolor{Gray} \textbf{5.92}
            & \cellcolor{Gray} \textbf{58.6}
            & \cellcolor{Gray} \textbf{64.5}
            & \cellcolor{Gray} \textbf{53.9}
            & \cellcolor{Gray} \textbf{64.5}
            & \cellcolor{Gray} \textbf{64.4} 
            
            \\
        & \cellcolor{Gray}\textbf{Ours 4D} 
			& \cellcolor{Gray} \textbf{7.18}
            & \cellcolor{Gray} \textbf{6.07}
            & \cellcolor{Gray} \textbf{4.44}
            & \cellcolor{Gray} \textbf{9.94}
            & \cellcolor{Gray} \textbf{6.59}
            & \cellcolor{Gray} \textbf{6.89}
            & \cellcolor{Gray} \textbf{5.28}
            & \cellcolor{Gray} \textbf{60.5}
            & \cellcolor{Gray} \textbf{65.7}
            & \cellcolor{Gray} \textbf{57.3}
            & \cellcolor{Gray} \textbf{65.7}
            & \cellcolor{Gray} \textbf{68.7}
            
            \\
        \midrule
        \multirow{7}{*}{\shortstack{2.25\\ \\W2\\g64}} & RTN 
            & 432 & 26.2 & 10.3 & - & - &  71.5 & 156 
            & 42.4  & 46.4 & - & 44.8 & 46.9 \\
         & GPTQ  
            & 20.9 & 22.4 & NAN & - & - &  14.2 & 10.1 
            & 47.5 & 54.2 &  -  & 51.8 & 48.8 \\
         & AWQ  
            & 2e5 & 1e5 & - & - & - & - & - & - & - & - & - & -\\
         & OQ    & \underline{9.62} & 7.56 & 6.11 & - & - & - & - & - & - & - & - & -\\ 
         & \cellcolor{Gray}\textbf{Ours 1D} 
			& \cellcolor{Gray} 10.1
            & \cellcolor{Gray} \textbf{6.99}
            & \cellcolor{Gray} \textbf{4.85}
            & \cellcolor{Gray} \textbf{14.1}
            & \cellcolor{Gray} \textbf{8.31}
            & \cellcolor{Gray} \textbf{9.69}
            & \cellcolor{Gray} \textbf{7.75}
            & \cellcolor{Gray} \textbf{52.8}
            & \cellcolor{Gray} \textbf{63.3}
            & \cellcolor{Gray} \textbf{57.3}
            & \cellcolor{Gray} \textbf{56.3}
            & \cellcolor{Gray} \textbf{57.4}
            
            \\
         & \cellcolor{Gray}\textbf{Ours 2D} 
			& \cellcolor{Gray} \textbf{7.61}
            & \cellcolor{Gray} \textbf{6.41}
            & \cellcolor{Gray} \textbf{4.58}
            & \cellcolor{Gray} \textbf{10.8}
            & \cellcolor{Gray} \textbf{6.83}
            & \cellcolor{Gray} \textbf{7.24}
            & \cellcolor{Gray} \textbf{5.58}
            & \cellcolor{Gray} \textbf{61.5}
            & \cellcolor{Gray} \textbf{64.8}
            & \cellcolor{Gray} \textbf{60.3}
            & \cellcolor{Gray} \textbf{65.3}
            & \cellcolor{Gray} \textbf{65.7}
            
            \\
        & \cellcolor{Gray}\textbf{Ours 4D} 
			& \cellcolor{Gray} \textbf{6.99}
            & \cellcolor{Gray} \textbf{5.98}
            & \cellcolor{Gray} \textbf{4.36}
            & \cellcolor{Gray} \textbf{9.59}
            & \cellcolor{Gray} \textbf{6.21}
            & \cellcolor{Gray} \textbf{6.66}
            & \cellcolor{Gray} \textbf{5.16}
            & \cellcolor{Gray} \textbf{62.9}
            & \cellcolor{Gray} \textbf{67.5}
            & \cellcolor{Gray} \textbf{62.3}
            & \cellcolor{Gray} \textbf{68.2}
            & \cellcolor{Gray} \textbf{69.3}
            
            \\
        \midrule
        \multirow{6}{*}{\shortstack{3.125\\ \\W3\\g128}} 
        & RTN 
            &  6.66 & 5.51 & 3.97 & 27.9 & 11.8 & 6.15 & 5.18 
            & 67.3 & 70.8 & 40.2 & 71.8 & 72.4  \\
        & GPTQ 
            & 6.29 & 5.42 & 3.85 & 8.2 & 5.2 & 5.83 & 4.71 
            & 66.2 & \underline{71.4} & 61.7  & \underline{72.2} & 72.7 \\
         & AWQ  & 6.24 & 5.32 & - & 8.2 & 4.8 & - & - & - & - & - & - & - \\
         & OQ  & 6.03 & 5.28 & 3.78 & - & - & - & - & - & - & - & - & - \\ 
          & \cellcolor{Gray}\textbf{Ours 1D} 
			& \cellcolor{Gray} \textbf{5.95}
            & \cellcolor{Gray} \textbf{5.19}
            & \cellcolor{Gray} \textbf{3.64}
            & \cellcolor{Gray} \textbf{7.29}
            & \cellcolor{Gray} \textbf{4.29}
            & \cellcolor{Gray} \textbf{5.79}
            & \cellcolor{Gray} \textbf{4.59}
            & \cellcolor{Gray} \textbf{66.9}
            & \cellcolor{Gray} 71.4
            & \cellcolor{Gray} \textbf{65.7}
            & \cellcolor{Gray} 71.0
            & \cellcolor{Gray} \textbf{73.5}
            
            \\
          & \cellcolor{Gray}\textbf{Ours 2D} 
			& \cellcolor{Gray} \textbf{5.83}
            & \cellcolor{Gray} \textbf{5.12}
            & \cellcolor{Gray} \textbf{3.58}
            & \cellcolor{Gray} \textbf{7.00}
            & \cellcolor{Gray} \textbf{4.04}
            & \cellcolor{Gray} \textbf{5.51}
            & \cellcolor{Gray} \textbf{4.27}
            & \cellcolor{Gray} \textbf{68.3}
            & \cellcolor{Gray} \textbf{71.2}
            & \cellcolor{Gray} \textbf{66.1}
            & \cellcolor{Gray} \textbf{73.9}
            & \cellcolor{Gray} \textbf{75.1}
            
            \\
        \bottomrule
    \end{tabular}
    \vspace{0.1cm}
    \label{tab:llama_weight_only}
\end{table*}
Table~\ref{tab:main_results} summarizes
results for GPTVQ, where 
we report WikiText 2 perplexity and an average over zero-shot task scores for the PIQA, BoolQ, ARC-easy, ARC-challenge, HellaSwag and WinoGrande tasks. 
We include
all Llama-v2 models, Mistral-7B-v0.1 and Mixtral-8x7B-v0.1.
More results are 
in appendix \ref{app:extra_results}:
Table~\ref{tab:lm_eval_results_llama} and Table~\ref{tab:lm_eval_results_mistral} contain individual scores for the zero-shot tasks, 
Table~\ref{tab:llama1_weight_only} contains WikiText2 perplexity for all Llama-v1 models, and Table~\ref{tab:extra_ppl_4bit} shows perplexity on 4 bit quantization.
Full VQ configurations can be found in Table~\ref{tab:configurations}.

These tables show
that non-uniform quantization using GPTVQ generally yields improved results over uniform PTQ methods.
This gap becomes especially large at low bitwidths and for very large models.
For example, comparing GPTVQ 2D on Llamav2-70B to OmniQuant W2@g128, we see an improvement of nearly 2 perplexity points.
Furthermore, in nearly all cases, 2D VQ outperforms 1D VQ, while 4D VQ shows even more significant improvements.

\paragraph{Comparison to other VQ methods}
\label{sec:other_vq}
Table~\ref{tab:other_vq} compares the results of GPTVQ to AQLM and QuIP\#~\cite{tseng2024quip}.
We use the same calibration data for AQLM and GPTVQ (SlimPajama, 4096 samples $\times$ 2048 tokens).
Furthermore, we use the the original AQLM source code and closely follow the AQLM evaluation protocol.
In particular, we use sequences of 4096 tokens from the WikiText2 test dataset, and average zero-shot accuracy among five LLM-evaluation-harness tasks (WinoGrande, PiQA, HellaSwag, ArcE, ArcC). 

In terms of size vs accuracy, GPTVQ is on par with AQLM when full block fine-tuning is not used.
Extra block fine-tuning allows AQLM to achieve better performance but increases the overall compression time significantly.
Note that GPTVQ was specifically designed for mobile CPU use case.
For this reason, we use smaller codebooks and a smaller vector dimension, even though this may lead to slightly worse perplexity and accuracy compared to the 8-dimensional case of AQLM and QuIP\#, c.f. Figure~\ref{fig:quant_sqnr}.
Further details and comparison can be found in Appendix~\ref{app:aqlm}.

\begin{table*}[ht!]
\centering
\footnotesize
\setlength{\tabcolsep}{5pt}
\caption{
\textbf{Comparison with existing SOTA VQ methods.}
No BFT option in AQLM denotes no block fine-tuning.
This setting is the most comparable to Our GPTVQ algorithm.
For fair compression comparison, we used the original AQLM code adapting it to our GPTVQ setup as close as possible, and also evluation protocol which is slightly different from Table~\ref{tab:llama_weight_only}.
}\label{tab:other_vq}
\vspace{0.2cm}
\begin{tabular}{llccccc}
    \toprule
    Model & Algorithm & bpv & Format & WikiText2 perplexity $\downarrow$ & Zero-shot avg. acc. $\uparrow$ & Compression time, h $\downarrow$ \\
    \midrule
    \multirow{4}{*}{L2-7B} & QuIP\# & 2.02 & 8D & 8.22 & 52.2 & - \\
    & AQLM (no BFT) & 2.29 & 8D & 7.49 & 58.9 & 18.3 \\
    & AQLM & 2.29 & 8D & \bf{6.65} & \bf{60.3} & 35.2 \\
    & \cellcolor{Gray} Ours & \cellcolor{Gray} 2.25 & \cellcolor{Gray} 4D & \cellcolor{Gray} 7.11 & \cellcolor{Gray} 59.2 & \cellcolor{Gray} \bf{2.5} \\
    \midrule
    \multirow{3}{*}{L3-8B} & AQLM (no BFT) & 2.27 & 8D & 9.86 & 60.9 & 19.9 \\
    & AQLM & 2.29 & 8D & \bf{8.71} & \bf{62.7} & 40.4 \\
    & \cellcolor{Gray} Ours & \cellcolor{Gray} 2.25 & \cellcolor{Gray} 4D & \cellcolor{Gray} 9.40 & \cellcolor{Gray} 60.0 & \cellcolor{Gray} \bf{2.8} \\
    \midrule
    \multirow{4}{*}{M-7B} & QuIP\# & 2.01 & 8D & 6.02 & 62.2 & - \\
    & AQLM (no BFT) & 2.27 & 8D & 6.87 & 64.0 & 19.4 \\
    & AQLM & 2.29 & 8D & \bf{5.77} & \bf{65.4} & 76.0 \\
    & \cellcolor{Gray} Ours & \cellcolor{Gray} 2.25 & \cellcolor{Gray} 4D & \cellcolor{Gray} 6.68 & \cellcolor{Gray} 62.7 & \cellcolor{Gray} \bf{2.8} \\
    \bottomrule
\end{tabular}
\end{table*}

\subsection{4 bit codebooks}
Reducing the bit width of the codebooks to 4 bit can bring further benefits to VQ implementations: 1) 2 values can be decoded with 1 LUT instruction, 2) codebook overhead is reduced, and 3) decoded tensors take up less space on on-chip cache memory. 
However, reducing the codebook bit width potentially lowers accuracy.
To mitigate the accuracy degradation we decode to groupwise INT4, with groups of 128 values. We refer to these groups as \textit{codebook quantization} groups, to distinguish them from the groups used in GPTVQ, which we denote \textit{GPTVQ groups}. We then modify the GPTVQ algorithm in the following ways: 1) First, for each group of weights $\textbf{w}_g$ we find the scale $s_g$ that minimizes $|(\textbf{w}_g - Q(\textbf{w}_g, s, b)) \textbf{h}_b|^2_2$, where $Q(\textbf{w}_g, s, b))$ denotes $b$-bit quantization of $\textbf{w}_g$, using scale $s$, and $\textbf{h}_g$ is the Hessian diagonal corresponding to block $\textbf{w}_g$. 2) Then, EM codebook initialization is applied to the GPTVQ group of weights. These groups are generally larger than the codebook quantization groupsize of 128. E.g., for $d=2, b=3$, a group size of $8192$ is used. Before EM, each quantzation group $g$ (i.e., row) in a block is scaled using the scale $s_g$ found in the previous step, but not clipped or rounded. This ensures that the codebook found is already on the INT4 scale, but that no further error is introduced. I.e., each row is scaled as $\textbf{w}_g^{s}=\frac{1}{s_g}\textbf{w}_g$, where the scaled rows $\textbf{w}_g^{s}$ are used in EM. 3) After EM, the codebook is clipped and rounded, but not scaled, since it is already scaled before initialization. 4) Lastly, during GPTVQ, each codebook quantization group is scaled using the scale found in step 1 before the nearest centroid is found: Equation~\ref{eq:kmeans_objective} is modified as $\textbf{c}^{(i)}=\arg\min_{m}  \left(\frac{1}{s^{(i)}}\textbf{x}^{(i)}-\textbf{c}^{(m)}\right)^T\textbf{D}^{(i)}\left(\frac{1}{s^{(i)}}\textbf{x}^{(i)}-\textbf{c}^{(m)}\right)$, where $s^{(i)}$ is the scale corresponding to vector $\textbf{x}^{(i)}$.

In Table~\ref{tab:int4_codebooks} we show the effect of quantizing codebooks to 4 bits. For Mistral-7B, a slightly modified version of the procedure enumerated above is used, in which the quantization group is set equal to the GPTVQ group. For all models considered, quantizing the codebook to INT4 yields very minor increase in perplexity.

\begin{table}
\centering
\vspace{-.25cm}
\normalsize
\setlength{\tabcolsep}{5pt}
\caption{\textbf{INT4 codebooks.} Perplexity for on WikiText2 for models with 2D, 3B VQ, blocksize 4096, with codebook entries quantized to INT4. INT4 CB column also shows PPL increase relative to INT8 CB. ${}^{*}$For Mistral-7B better results were achieved by avoiding block quantization and instead quantizing per VQ block. }\label{tab:int4_codebooks}
\vspace{0.2cm}
\begin{tabular}{cccc}
    \toprule
    Model & INT4 b128 & INT8 CB & INT4 CB \\ 
    \midrule
    L2-7B & 5.72 & 5.95 & 5.99 (+0.05) \\
    L3-8B & 6.73 & 7.00 & 7.29 (+0.29) \\
    M-7B & 5.34 & 5.51 & 5.77${}^{*}$ (+0.26) \\
    \bottomrule
\end{tabular}
\end{table}

\subsection{Combination of GPTVQ with LoRA adapters}
Table~\ref{tab:lora} shows the results of fine-tuning LoRA adapters on top of GPTVQ representation.
We do not present the WikiText2 perplexity from the LoftQ paper because it was calculated using a different protocol than the one currently adopted in quantization literature~\cite{shao2023omniquant,egiazarian2024aqlm}. 
First, we observe that LoRA adapters not only help to improve WikiText2 perplexity and GSM8k~\cite{cobbe2021training} accuracy, but also helps to recover performance from aggressive quantization.
Second, we note that the GPTVQ and LoRA combination achieves a noteworthy improvement over previous methods making it a top choice for on device LLM serving with adapters, e.g.,~\cite{gunter2024appleintelligencefoundationlanguage}.

\begin{table*}[ht!]
\centering
\footnotesize
\setlength{\tabcolsep}{5pt}
\caption{
\textbf{Results of fine-tuning LoRA adapters.}
We fine-tuned LoRA adapters on WikiText2 train split for WikiText2 task, and on GSM8k train split for GSM8k task.
Subsequent evaluation was done on the corresponding test splits.
Frozen adapter is taken from LoRA with FP16 base model (second row) and applied to the vector-quantized base models.
We added 0.127 bpv overhead to LoftQ~\cite{li2023loftq} model mentioned in QLoRA~\cite{dettmers2024qlora}. 
N.a. indicates that the training diverged.
}
\label{tab:lora}
\vspace{0.2cm}
\begin{tabular}{lccccccccc}
    \toprule
    & & & & \multicolumn{3}{c}{WikiText2 perplexity $\downarrow$} & \multicolumn{3}{c}{GSM8k accuracy $\uparrow$} \\
    Method & Base bpv & Adapter & Format & L2-7B & L3-8B &  M-7B & L2-7B & L3-8B & M-7B \\ 
    \midrule
    FP & 16 & - & - & 5.47 & 6.14 & 5.25 & 2.4 & 6.5 & 16.4 \\
    LoRA & 16 &  Trained & - & 4.91 & 5.82 & 5.19 & 40.3 & 62.3 & 56.5 \\
    \midrule
    QLoRA & 2.127 & Trained & NF2 & - & - & - & n.a. & - & - \\
    LoftQ & 2.127 & Trained & NF2 & - & - & - & 20.9 & - & - \\
    \cellcolor{Gray} Ours & \cellcolor{Gray} 2.125 & \cellcolor{Gray} Frozen & \cellcolor{Gray} VQ 2D & \cellcolor{Gray} 7.18 & \cellcolor{Gray} 10.73 & \cellcolor{Gray} 7.25 & \cellcolor{Gray} 15.5 & \cellcolor{Gray} 19.3 & \cellcolor{Gray} 33.4 \\
    \cellcolor{Gray} Ours & \cellcolor{Gray} 2.125 & \cellcolor{Gray} Trained & \cellcolor{Gray} VQ 2D & \cellcolor{Gray} \bf{6.04} & \cellcolor{Gray} \bf{8.52} & \cellcolor{Gray} \bf{6.19} & \cellcolor{Gray} \bf{33.4} & \cellcolor{Gray} \bf{50.0} & \cellcolor{Gray} \bf{50.2} \\
    \midrule
    \cellcolor{Gray} Ours & \cellcolor{Gray} 2.125 & \cellcolor{Gray} Frozen & \cellcolor{Gray} VQ 4D & \cellcolor{Gray} 6.53 & \cellcolor{Gray} 9.65 & \cellcolor{Gray} 6.55 & \cellcolor{Gray} 18.2 & \cellcolor{Gray} 29.9 & \cellcolor{Gray} 39.6 \\
    \cellcolor{Gray} Ours & \cellcolor{Gray} 2.125 & \cellcolor{Gray} Trained & \cellcolor{Gray} VQ 4D & \cellcolor{Gray} \bf{5.83} & \cellcolor{Gray} \bf{8.07} & \cellcolor{Gray} \bf{6.00} & \cellcolor{Gray} \bf{32.5} & \cellcolor{Gray} \bf{51.0} & \cellcolor{Gray} \bf{51.8} \\
    \midrule
    \cellcolor{Gray} Ours & \cellcolor{Gray} 2.25 & \cellcolor{Gray} Frozen & \cellcolor{Gray} VQ 2D & \cellcolor{Gray} 6.99 & \cellcolor{Gray} 10.23 & \cellcolor{Gray} 6.97 & \cellcolor{Gray} 17.1 & \cellcolor{Gray} 24.8 & \cellcolor{Gray} 35.4 \\
    \cellcolor{Gray} Ours & \cellcolor{Gray} 2.25 & \cellcolor{Gray} Trained & \cellcolor{Gray} VQ 2D & \cellcolor{Gray} \bf{5.97} & \cellcolor{Gray} \bf{8.32} & \cellcolor{Gray} \bf{6.13} & \cellcolor{Gray} \bf{34.8} & \cellcolor{Gray} \bf{50.2} & \cellcolor{Gray} \bf{49.3} \\
    \midrule
    \cellcolor{Gray} Ours & \cellcolor{Gray} 2.25 & \cellcolor{Gray} Frozen & \cellcolor{Gray} VQ 4D & \cellcolor{Gray} 6.27 & \cellcolor{Gray} 9.13 & \cellcolor{Gray} 6.36 & \cellcolor{Gray} 19.9 & \cellcolor{Gray} 32.2 & \cellcolor{Gray} 42.8 \\
    \cellcolor{Gray} Ours & \cellcolor{Gray} 2.25 & \cellcolor{Gray} Trained & \cellcolor{Gray} VQ 4D & \cellcolor{Gray} \bf{5.77} & \cellcolor{Gray} \bf{7.94} & \cellcolor{Gray} \bf{5.94} & \cellcolor{Gray} \bf{35.0} & \cellcolor{Gray} \bf{52.5} & \cellcolor{Gray} \bf{50.4} \\
    \midrule
    QLoRA & 3.127 & Trained & NF2/NF4 & - & - & - & 32.1 & - & - \\
    LoftQ & 3.127 & Trained & NF2/NF4 & - & - & - & 32.9 & - & - \\
    \cellcolor{Gray} Ours & \cellcolor{Gray} 3.125 & \cellcolor{Gray} Frozen & \cellcolor{Gray} VQ 2D & \cellcolor{Gray} 5.25 & \cellcolor{Gray} 6.71 & \cellcolor{Gray} 5.46 & \cellcolor{Gray} 34.8 & \cellcolor{Gray} 53.9 & \cellcolor{Gray} 52.3 \\
    \cellcolor{Gray} Ours & \cellcolor{Gray} 3.125 & \cellcolor{Gray} Trained & \cellcolor{Gray} VQ 2D & \cellcolor{Gray} \bf{5.18} & \cellcolor{Gray} \bf{6.62} & \cellcolor{Gray} \bf{5.42} & \cellcolor{Gray} \bf{37.2} & \cellcolor{Gray} \bf{55.4} & \cellcolor{Gray} \bf{57.0} \\
    \bottomrule
\end{tabular}
\end{table*}

\section{Related work}
\label{sec:related_work}
\paragraph{Vector quantization}
A number of works propose vector quantization of CNN weights~\cite{gong2014compressing, martinez2021permute, fan2020training, stock2019and, wu2016quantized, martinez2021permute, cho2021dkm}. The most common approach is to reshape the weights of convolutional or fully connected layers into a matrix, and then apply K-means clustering directly on the columns. Typically, the clustering is applied on scalars or vectors of dimension 4 or higher. Some of these works consider data-aware optimization of the quantized weights. Most often, a variant of the EM algorithm is used in order to update centroids and indices~\cite{stock2019and, gong2014compressing}. An alternative approach is using a differentiable K-means formulation, which enables fine-tuning using SGD with the original loss function in order to recover the network accuracy~\cite{cho2021dkm, fan2020training, tang2023lut}.
Finally, most recent SOTA works~\cite{tseng2024quip,egiazarian2024aqlm,malinovskii2024pvtuning} apply layer-wise, block-wise and even end-to-end fine-tuning on a calibration dataset to recover accuracy loss.

\vspace{-0.2cm}
\paragraph{LLM quantization} Applying DNN quantization approaches to recent LLMs often poses significant computational challenges. Therefore, even uniform post-training quantization methods 
must be optimized for
scalability~\cite{frantar2022gptq}. Since vector quantization 
approaches have even higher computational complexity, 
applying them to LLM weights compression may be expensive.
The most similar to our work is 
a method~\cite{deng2024llmcodebook}, which 
uses gradient-based layer sensitivities to update the codebooks and a reduced complexity LoRA-based approach~\cite{hu2021lora} to partially recover the accuracy.

\vspace{-0.2cm}
\paragraph{Hessian-based compression methods} Several classical works suggest second-order approximation of the neural network loss function for accurate unstructured pruning~\cite{lecun1989optimal, hassibi1993optimal}. A more recent line of work extends this family of methods to PTQ~\cite{singh2020woodfisher,frantar2022optimal, frantar2022gptq}.

\section{Conclusions}
In this work, we have shown that vector quantization, when properly designed, can yield significant positive impact on token generation speed, DRAM footprint, and perplexity on a mobile CPU.
Furthermore, we have shown that VQ in one or more dimensions progressively improves quantized large language model accuracy. 
By co-designing the VQ representation with the mobile CPU implementation, we created a fast decode kernel, leveraging existing hardware LUT instructions, and demonstrated an increase in decode token rate.
Lastly, we introduced GPTVQ, a fast method for post-training quantization of LLMs that achieves competitive model size vs accuracy trade-offs on a wide range of LLMs and zero-shot tasks, using mobile friendly VQ configurations.

\bibliography{references}

\begin{thebibliography}{52}
\providecommand{\natexlab}[1]{#1}
\providecommand{\url}[1]{\texttt{#1}}
\expandafter\ifx\csname urlstyle\endcsname\relax
  \providecommand{\doi}[1]{doi: #1}\else
  \providecommand{\doi}{doi: \begingroup \urlstyle{rm}\Url}\fi

\bibitem[Arthur \& Vassilvitskii(2007)Arthur and Vassilvitskii]{arthur2007k}
Arthur, D. and Vassilvitskii, S.
\newblock K-means++ the advantages of careful seeding.
\newblock In \emph{Proceedings of the eighteenth annual ACM-SIAM symposium on Discrete algorithms}, pp.\  1027--1035, 2007.

\bibitem[Bishop(2006)]{bishop2006pattern}
Bishop, C.~M.
\newblock Pattern recognition and machine learning.
\newblock \emph{Springer google schola}, 2:\penalty0 1122--1128, 2006.

\bibitem[Bisk et~al.(2020)Bisk, Zellers, Gao, Choi, et~al.]{bisk2020piqa}
Bisk, Y., Zellers, R., Gao, J., Choi, Y., et~al.
\newblock Piqa: Reasoning about physical commonsense in natural language.
\newblock In \emph{Proceedings of the AAAI conference on artificial intelligence}, volume~34, pp.\  7432--7439, 2020.

\bibitem[Cheng et~al.(2024)Cheng, Leng, Zhang, Xin, Li, Chen, Zhu, Zhang, Luo, Zhao, and Bing]{damonlpsg2024videollama2}
Cheng, Z., Leng, S., Zhang, H., Xin, Y., Li, X., Chen, G., Zhu, Y., Zhang, W., Luo, Z., Zhao, D., and Bing, L.
\newblock Videollama 2: Advancing spatial-temporal modeling and audio understanding in video-llms.
\newblock \emph{arXiv preprint arXiv:2406.07476}, 2024.

\bibitem[Cho et~al.(2021)Cho, Vahid, Adya, and Rastegari]{cho2021dkm}
Cho, M., Vahid, K.~A., Adya, S., and Rastegari, M.
\newblock Dkm: Differentiable k-means clustering layer for neural network compression.
\newblock \emph{arXiv preprint arXiv:2108.12659}, 2021.

\bibitem[Clark et~al.(2019)Clark, Lee, Chang, Kwiatkowski, Collins, and Toutanova]{clark2019boolq}
Clark, C., Lee, K., Chang, M.-W., Kwiatkowski, T., Collins, M., and Toutanova, K.
\newblock Boolq: Exploring the surprising difficulty of natural yes/no questions.
\newblock In \emph{NAACL}, 2019.

\bibitem[Clark et~al.(2018)Clark, Cowhey, Etzioni, Khot, Sabharwal, Schoenick, and Tafjord]{allenai:arc}
Clark, P., Cowhey, I., Etzioni, O., Khot, T., Sabharwal, A., Schoenick, C., and Tafjord, O.
\newblock Think you have solved question answering? try arc, the ai2 reasoning challenge.
\newblock \emph{arXiv:1803.05457v1}, 2018.

\bibitem[Cobbe et~al.(2021)Cobbe, Kosaraju, Bavarian, Chen, Jun, Kaiser, Plappert, Tworek, Hilton, Nakano, et~al.]{cobbe2021training}
Cobbe, K., Kosaraju, V., Bavarian, M., Chen, M., Jun, H., Kaiser, L., Plappert, M., Tworek, J., Hilton, J., Nakano, R., et~al.
\newblock Training verifiers to solve math word problems.
\newblock \emph{arXiv preprint arXiv:2110.14168}, 2021.

\bibitem[Deng et~al.(2024)Deng, Li, Wang, Gu, Shen, and Huang]{deng2024llmcodebook}
Deng, J., Li, S., Wang, C., Gu, H., Shen, H., and Huang, K.
\newblock {LLM}-codebook for extreme compression of large language models, 2024.

\bibitem[Dettmers et~al.(2024)Dettmers, Pagnoni, Holtzman, and Zettlemoyer]{dettmers2024qlora}
Dettmers, T., Pagnoni, A., Holtzman, A., and Zettlemoyer, L.
\newblock Qlora: Efficient finetuning of quantized llms.
\newblock \emph{Advances in Neural Information Processing Systems}, 36, 2024.

\bibitem[Egiazarian et~al.(2024)Egiazarian, Panferov, Kuznedelev, Frantar, Babenko, and Alistarh]{egiazarian2024aqlm}
Egiazarian, V., Panferov, A., Kuznedelev, D., Frantar, E., Babenko, A., and Alistarh, D.
\newblock Extreme compression of large language models via additive quantization.
\newblock \emph{arXiv preprint arXiv:2401.06118}, 2024.

\bibitem[Fan et~al.(2020)Fan, Stock, Graham, Grave, Gribonval, Jegou, and Joulin]{fan2020training}
Fan, A., Stock, P., Graham, B., Grave, E., Gribonval, R., Jegou, H., and Joulin, A.
\newblock Training with quantization noise for extreme model compression.
\newblock \emph{arXiv preprint arXiv:2004.07320}, 2020.

\bibitem[Frantar \& Alistarh(2022)Frantar and Alistarh]{frantar2022optimal}
Frantar, E. and Alistarh, D.
\newblock Optimal brain compression: A framework for accurate post-training quantization and pruning.
\newblock \emph{Advances in Neural Information Processing Systems}, 35:\penalty0 4475--4488, 2022.

\bibitem[Frantar et~al.(2022)Frantar, Ashkboos, Hoefler, and Alistarh]{frantar2022gptq}
Frantar, E., Ashkboos, S., Hoefler, T., and Alistarh, D.
\newblock Gptq: Accurate post-training quantization for generative pre-trained transformers.
\newblock \emph{arXiv preprint arXiv:2210.17323}, 2022.

\bibitem[Gao et~al.(2023)Gao, Tow, Abbasi, Biderman, Black, DiPofi, Foster, Golding, Hsu, Le~Noac'h, Li, McDonell, Muennighoff, Ociepa, Phang, Reynolds, Schoelkopf, Skowron, Sutawika, Tang, Thite, Wang, Wang, and Zou]{eval-harness}
Gao, L., Tow, J., Abbasi, B., Biderman, S., Black, S., DiPofi, A., Foster, C., Golding, L., Hsu, J., Le~Noac'h, A., Li, H., McDonell, K., Muennighoff, N., Ociepa, C., Phang, J., Reynolds, L., Schoelkopf, H., Skowron, A., Sutawika, L., Tang, E., Thite, A., Wang, B., Wang, K., and Zou, A.
\newblock A framework for few-shot language model evaluation, 12 2023.
\newblock URL \url{https://zenodo.org/records/10256836}.

\bibitem[Gersho \& Gray(2012)Gersho and Gray]{gersho2012vector}
Gersho, A. and Gray, R.~M.
\newblock \emph{Vector quantization and signal compression}, volume 159.
\newblock Springer Science \& Business Media, 2012.

\bibitem[Gholami et~al.(2022)Gholami, Kim, Dong, Yao, Mahoney, and Keutzer]{gholami2022survey}
Gholami, A., Kim, S., Dong, Z., Yao, Z., Mahoney, M.~W., and Keutzer, K.
\newblock A survey of quantization methods for efficient neural network inference.
\newblock In \emph{Low-Power Computer Vision}, pp.\  291--326. Chapman and Hall/CRC, 2022.

\bibitem[Gong et~al.(2014)Gong, Liu, Yang, and Bourdev]{gong2014compressing}
Gong, Y., Liu, L., Yang, M., and Bourdev, L.
\newblock Compressing deep convolutional networks using vector quantization.
\newblock \emph{arXiv preprint arXiv:1412.6115}, 2014.

\bibitem[Gunter et~al.(2024)Gunter, Wang, Wang, Pang, Narayanan, Zhang, Zhang, Chen, Chiu, Qiu, Gopinath, Yap, Yin, Nan, Weers, Yin, Huang, Wang, Lu, Peebles, Ye, Lee, Du, Chen, Keunebroek, Wiseman, Evans, Lei, Rathod, Kong, Du, Li, Wang, Gao, Ahmed, Xu, Lu, Rashid, Jose, Doane, Bencomo, Vanderby, Hansen, Jain, Anupama, Kamal, Wu, Brum, Maalouf, Erdenebileg, Dulhanty, Moritz, Kang, Jimenez, Ladd, Shi, Bai, Chu, Hohman, Kotek, Coleman, Li, Bigham, Cao, Lai, Cheung, Shan, Zhou, Li, Qin, Singh, Vega, Zou, Heckman, Gardiner, Bowler, Cordell, Cao, Hay, Shahdadpuri, Godwin, Dighe, Rachapudi, Tantawi, Frigg, Davarnia, Shah, Guha, Sirovica, Ma, Ma, Wang, Kim, Jayaram, Shankar, Paidi, Kumar, Wang, Zheng, Cheng, Shrager, Ye, Tanaka, Guo, Meng, Luo, Ouyang, Aygar, Wan, Walkingshaw, Narayanan, Lin, Farooq, Ramerth, Reed, Bartels, Chaney, Riazati, Yang, Feldman, Hochstrasser, Seguin, Belousova, Pelemans, Yang, Vahid, Cao, Najibi, Zuliani, Horton, Cho, Bhendawade, Dong, Maj, Agrawal, Shan, Fu, Poston, Xu, Liu, Rao,
  Heeramun, Merth, Rayala, Cui, Sridhar, Zhang, Zhang, Wu, Zhou, Liu, Zhao, Xia, Ren, and Ren]{gunter2024appleintelligencefoundationlanguage}
Gunter, T., Wang, Z., Wang, C., Pang, R., Narayanan, A., Zhang, A., Zhang, B., Chen, C., Chiu, C.-C., Qiu, D., Gopinath, D., Yap, D.~A., Yin, D., Nan, F., Weers, F., Yin, G., Huang, H., Wang, J., Lu, J., Peebles, J., Ye, K., Lee, M., Du, N., Chen, Q., Keunebroek, Q., Wiseman, S., Evans, S., Lei, T., Rathod, V., Kong, X., Du, X., Li, Y., Wang, Y., Gao, Y., Ahmed, Z., Xu, Z., Lu, Z., Rashid, A., Jose, A.~M., Doane, A., Bencomo, A., Vanderby, A., Hansen, A., Jain, A., Anupama, A.~M., Kamal, A., Wu, B., Brum, C., Maalouf, C., Erdenebileg, C., Dulhanty, C., Moritz, D., Kang, D., Jimenez, E., Ladd, E., Shi, F., Bai, F., Chu, F., Hohman, F., Kotek, H., Coleman, H.~G., Li, J., Bigham, J., Cao, J., Lai, J., Cheung, J., Shan, J., Zhou, J., Li, J., Qin, J., Singh, K., Vega, K., Zou, K., Heckman, L., Gardiner, L., Bowler, M., Cordell, M., Cao, M., Hay, N., Shahdadpuri, N., Godwin, O., Dighe, P., Rachapudi, P., Tantawi, R., Frigg, R., Davarnia, S., Shah, S., Guha, S., Sirovica, S., Ma, S., Ma, S., Wang, S., Kim, S.,
  Jayaram, S., Shankar, V., Paidi, V., Kumar, V., Wang, X., Zheng, X., Cheng, W., Shrager, Y., Ye, Y., Tanaka, Y., Guo, Y., Meng, Y., Luo, Z.~T., Ouyang, Z., Aygar, A., Wan, A., Walkingshaw, A., Narayanan, A., Lin, A., Farooq, A., Ramerth, B., Reed, C., Bartels, C., Chaney, C., Riazati, D., Yang, E.~L., Feldman, E., Hochstrasser, G., Seguin, G., Belousova, I., Pelemans, J., Yang, K., Vahid, K.~A., Cao, L., Najibi, M., Zuliani, M., Horton, M., Cho, M., Bhendawade, N., Dong, P., Maj, P., Agrawal, P., Shan, Q., Fu, Q., Poston, R., Xu, S., Liu, S., Rao, S., Heeramun, T., Merth, T., Rayala, U., Cui, V., Sridhar, V.~R., Zhang, W., Zhang, W., Wu, W., Zhou, X., Liu, X., Zhao, Y., Xia, Y., Ren, Z., and Ren, Z.
\newblock Apple intelligence foundation language models, 2024.

\bibitem[Hassibi et~al.(1993)Hassibi, Stork, and Wolff]{hassibi1993optimal}
Hassibi, B., Stork, D.~G., and Wolff, G.~J.
\newblock Optimal brain surgeon and general network pruning.
\newblock In \emph{IEEE international conference on neural networks}, pp.\  293--299. IEEE, 1993.

\bibitem[Hooper et~al.(2024)Hooper, Kim, Mohammadzadeh, Mahoney, Shao, Keutzer, and Gholami]{hooper2024kvquant}
Hooper, C., Kim, S., Mohammadzadeh, H., Mahoney, M.~W., Shao, Y.~S., Keutzer, K., and Gholami, A.
\newblock Kvquant: Towards 10 million context length llm inference with kv cache quantization.
\newblock \emph{arXiv preprint arXiv:2401.18079}, 2024.

\bibitem[Hu et~al.(2021)Hu, Shen, Wallis, Allen-Zhu, Li, Wang, Wang, and Chen]{hu2021lora}
Hu, E.~J., Shen, Y., Wallis, P., Allen-Zhu, Z., Li, Y., Wang, S., Wang, L., and Chen, W.
\newblock Lora: Low-rank adaptation of large language models.
\newblock \emph{arXiv preprint arXiv:2106.09685}, 2021.

\bibitem[Huang et~al.(2024)Huang, Ma, Qin, Zheng, Lv, Chen, Luo, Qi, Liu, and Magno]{huang2024llama3quant}
Huang, W., Ma, X., Qin, H., Zheng, X., Lv, C., Chen, H., Luo, J., Qi, X., Liu, X., and Magno, M.
\newblock How good are low-bit quantized {L}lama3 models? {A}n empirical study.
\newblock \emph{arXiv preprint arXiv:2404.14047}, 2024.

\bibitem[Jiang et~al.(2023)Jiang, Sablayrolles, Mensch, Bamford, Chaplot, Casas, Bressand, Lengyel, Lample, Saulnier, et~al.]{jiang2023mistral}
Jiang, A.~Q., Sablayrolles, A., Mensch, A., Bamford, C., Chaplot, D.~S., Casas, D. d.~l., Bressand, F., Lengyel, G., Lample, G., Saulnier, L., et~al.
\newblock Mistral 7b.
\newblock \emph{arXiv preprint arXiv:2310.06825}, 2023.

\bibitem[Jiang et~al.(2024)Jiang, Sablayrolles, Roux, Mensch, Savary, Bamford, Chaplot, Casas, Hanna, Bressand, et~al.]{jiang2024mixtral}
Jiang, A.~Q., Sablayrolles, A., Roux, A., Mensch, A., Savary, B., Bamford, C., Chaplot, D.~S., Casas, D. d.~l., Hanna, E.~B., Bressand, F., et~al.
\newblock Mixtral of experts.
\newblock \emph{arXiv preprint arXiv:2401.04088}, 2024.

\bibitem[Kim et~al.(2023)Kim, Hooper, Gholami, Dong, Li, Shen, Mahoney, and Keutzer]{kim2023squeezellm}
Kim, S., Hooper, C., Gholami, A., Dong, Z., Li, X., Shen, S., Mahoney, M.~W., and Keutzer, K.
\newblock Squeezellm: Dense-and-sparse quantization.
\newblock \emph{arXiv preprint arXiv:2306.07629}, 2023.

\bibitem[LeCun et~al.(1989)LeCun, Denker, and Solla]{lecun1989optimal}
LeCun, Y., Denker, J., and Solla, S.
\newblock Optimal brain damage.
\newblock \emph{Advances in neural information processing systems}, 2, 1989.

\bibitem[Li et~al.(2024)Li, Yu, Liang, He, Karampatziakis, Chen, and Zhao]{li2023loftq}
Li, Y., Yu, Y., Liang, C., He, P., Karampatziakis, N., Chen, W., and Zhao, T.
\newblock Loftq: Lora-fine-tuning-aware quantization for large language models.
\newblock In \emph{International Conference on Learning Representations}, 2024.

\bibitem[Lin et~al.(2023{\natexlab{a}})Lin, Zhu, Ye, Ning, Jin, and Yuan]{lin2023video}
Lin, B., Zhu, B., Ye, Y., Ning, M., Jin, P., and Yuan, L.
\newblock Video-llava: Learning united visual representation by alignment before projection.
\newblock \emph{arXiv preprint arXiv:2311.10122}, 2023{\natexlab{a}}.

\bibitem[Lin et~al.(2023{\natexlab{b}})Lin, Tang, Tang, Yang, Dang, and Han]{lin2023awq}
Lin, J., Tang, J., Tang, H., Yang, S., Dang, X., and Han, S.
\newblock Awq: Activation-aware weight quantization for llm compression and acceleration.
\newblock \emph{arXiv preprint arXiv:2306.00978}, 2023{\natexlab{b}}.

\bibitem[Lin et~al.(2023{\natexlab{c}})Lin, Yin, Ping, Lu, Molchanov, Tao, Mao, Kautz, Shoeybi, and Han]{lin2023vila}
Lin, J., Yin, H., Ping, W., Lu, Y., Molchanov, P., Tao, A., Mao, H., Kautz, J., Shoeybi, M., and Han, S.
\newblock Vila: On pre-training for visual language models, 2023{\natexlab{c}}.

\bibitem[Liu et~al.(2023)Liu, Li, Wu, and Lee]{liu2023llava}
Liu, H., Li, C., Wu, Q., and Lee, Y.~J.
\newblock Visual instruction tuning.
\newblock In \emph{NeurIPS}, 2023.

\bibitem[Malinovskii et~al.(2024{\natexlab{a}})Malinovskii, Mazur, Ilin, Kuznedelev, Burlachenko, Yi, Alistarh, and Richtarik]{malinovskii2024pv}
Malinovskii, V., Mazur, D., Ilin, I., Kuznedelev, D., Burlachenko, K., Yi, K., Alistarh, D., and Richtarik, P.
\newblock Pv-tuning: Beyond straight-through estimation for extreme llm compression.
\newblock \emph{Advances in Neural Information Processing Systems}, 37:\penalty0 5074--5121, 2024{\natexlab{a}}.

\bibitem[Malinovskii et~al.(2024{\natexlab{b}})Malinovskii, Mazur, Ilin, Kuznedelev, Burlachenko, Yi, Alistarh, and Richtarik]{malinovskii2024pvtuning}
Malinovskii, V., Mazur, D., Ilin, I., Kuznedelev, D., Burlachenko, K., Yi, K., Alistarh, D., and Richtarik, P.
\newblock Pv-tuning: Beyond straight-through estimation for extreme llm compression, 2024{\natexlab{b}}.

\bibitem[Martinez et~al.(2021)Martinez, Shewakramani, Liu, B{\^a}rsan, Zeng, and Urtasun]{martinez2021permute}
Martinez, J., Shewakramani, J., Liu, T.~W., B{\^a}rsan, I.~A., Zeng, W., and Urtasun, R.
\newblock Permute, quantize, and fine-tune: Efficient compression of neural networks.
\newblock In \emph{Proceedings of the IEEE/CVF Conference on Computer Vision and Pattern Recognition}, pp.\  15699--15708, 2021.

\bibitem[Merity et~al.(2016)Merity, Xiong, Bradbury, and Socher]{merity2016pointer}
Merity, S., Xiong, C., Bradbury, J., and Socher, R.
\newblock Pointer sentinel mixture models, 2016.

\bibitem[Nagel et~al.(2020)Nagel, Amjad, Van~Baalen, Louizos, and Blankevoort]{nagel2020adaround}
Nagel, M., Amjad, R.~A., Van~Baalen, M., Louizos, C., and Blankevoort, T.
\newblock Up or down? adaptive rounding for post-training quantization.
\newblock In \emph{International Conference on Machine Learning}, pp.\  7197--7206. PMLR, 2020.

\bibitem[Nagel et~al.(2021)Nagel, Fournarakis, Amjad, Bondarenko, Van~Baalen, and Blankevoort]{nagel2021whitepaper}
Nagel, M., Fournarakis, M., Amjad, R.~A., Bondarenko, Y., Van~Baalen, M., and Blankevoort, T.
\newblock A white paper on neural network quantization.
\newblock \emph{arXiv preprint arXiv:2106.08295}, 2021.

\bibitem[Sakaguchi et~al.(2021)Sakaguchi, Bras, Bhagavatula, and Choi]{ai2:winogrande}
Sakaguchi, K., Bras, R.~L., Bhagavatula, C., and Choi, Y.
\newblock Winogrande: An adversarial winograd schema challenge at scale.
\newblock \emph{Communications of the ACM}, 64\penalty0 (9):\penalty0 99--106, 2021.

\bibitem[Shao et~al.(2023)Shao, Chen, Zhang, Xu, Zhao, Li, Zhang, Gao, Qiao, and Luo]{shao2023omniquant}
Shao, W., Chen, M., Zhang, Z., Xu, P., Zhao, L., Li, Z., Zhang, K., Gao, P., Qiao, Y., and Luo, P.
\newblock Omniquant: Omnidirectionally calibrated quantization for large language models.
\newblock \emph{arXiv preprint arXiv:2308.13137}, 2023.

\bibitem[Singh \& Alistarh(2020)Singh and Alistarh]{singh2020woodfisher}
Singh, S.~P. and Alistarh, D.
\newblock Woodfisher: Efficient second-order approximation for neural network compression.
\newblock \emph{Advances in Neural Information Processing Systems}, 33:\penalty0 18098--18109, 2020.

\bibitem[Soboleva et~al.(2023)Soboleva, Al-Khateeb, Myers, Steeves, Hestness, and Dey]{cerebras2023slimpajama}
Soboleva, D., Al-Khateeb, F., Myers, R., Steeves, J.~R., Hestness, J., and Dey, N.
\newblock {SlimPajama: A 627B token cleaned and deduplicated version of RedPajama}, 2023.

\bibitem[Stock et~al.(2019)Stock, Joulin, Gribonval, Graham, and J{\'e}gou]{stock2019and}
Stock, P., Joulin, A., Gribonval, R., Graham, B., and J{\'e}gou, H.
\newblock And the bit goes down: Revisiting the quantization of neural networks.
\newblock \emph{arXiv preprint arXiv:1907.05686}, 2019.

\bibitem[Tang et~al.(2023)Tang, Wang, Cao, Zhang, Chen, Cai, Liu, and Yang]{tang2023lut}
Tang, X., Wang, Y., Cao, T., Zhang, L.~L., Chen, Q., Cai, D., Liu, Y., and Yang, M.
\newblock Lut-nn: Empower efficient neural network inference with centroid learning and table lookup.
\newblock In \emph{Proceedings of the 29th Annual International Conference on Mobile Computing and Networking}, pp.\  1--15, 2023.

\bibitem[Touvron et~al.(2023{\natexlab{a}})Touvron, Lavril, Izacard, Martinet, Lachaux, Lacroix, Rozi{\`e}re, Goyal, Hambro, Azhar, et~al.]{touvron2023llama}
Touvron, H., Lavril, T., Izacard, G., Martinet, X., Lachaux, M.-A., Lacroix, T., Rozi{\`e}re, B., Goyal, N., Hambro, E., Azhar, F., et~al.
\newblock Llama: Open and efficient foundation language models.
\newblock \emph{arXiv preprint arXiv:2302.13971}, 2023{\natexlab{a}}.

\bibitem[Touvron et~al.(2023{\natexlab{b}})Touvron, Martin, Stone, Albert, Almahairi, Babaei, Bashlykov, Batra, Bhargava, Bhosale, et~al.]{touvron2023llama2}
Touvron, H., Martin, L., Stone, K., Albert, P., Almahairi, A., Babaei, Y., Bashlykov, N., Batra, S., Bhargava, P., Bhosale, S., et~al.
\newblock Llama 2: Open foundation and fine-tuned chat models.
\newblock \emph{arXiv preprint arXiv:2307.09288}, 2023{\natexlab{b}}.

\bibitem[Tseng et~al.(2024)Tseng, Chee, Sun, Kuleshov, and Sa]{tseng2024quip}
Tseng, A., Chee, J., Sun, Q., Kuleshov, V., and Sa, C.~D.
\newblock Qu{IP}\#: Even better {LLM} quantization with hadamard incoherence and lattice codebooks.
\newblock In \emph{Forty-first International Conference on Machine Learning}, 2024.

\bibitem[{Wikipedia}(2024)]{wiki:iphone}
{Wikipedia}.
\newblock i{P}hone 16, 2024.
\newblock URL \url{https://en.wikipedia.org/wiki/IPhone_16}.

\bibitem[Workshop et~al.(2022)Workshop, Scao, Fan, Akiki, Pavlick, Ili{\'c}, Hesslow, Castagn{\'e}, Luccioni, Yvon, et~al.]{workshop2022bloom}
Workshop, B., Scao, T.~L., Fan, A., Akiki, C., Pavlick, E., Ili{\'c}, S., Hesslow, D., Castagn{\'e}, R., Luccioni, A.~S., Yvon, F., et~al.
\newblock Bloom: A 176b-parameter open-access multilingual language model.
\newblock \emph{arXiv preprint arXiv:2211.05100}, 2022.

\bibitem[Wu et~al.(2016)Wu, Leng, Wang, Hu, and Cheng]{wu2016quantized}
Wu, J., Leng, C., Wang, Y., Hu, Q., and Cheng, J.
\newblock Quantized convolutional neural networks for mobile devices.
\newblock In \emph{Proceedings of the IEEE Conference on Computer Vision and Pattern Recognition}, pp.\  4820--4828, 2016.

\bibitem[Zellers et~al.(2019)Zellers, Holtzman, Bisk, Farhadi, and Choi]{zellers2019hellaswag}
Zellers, R., Holtzman, A., Bisk, Y., Farhadi, A., and Choi, Y.
\newblock Hellaswag: Can a machine really finish your sentence?
\newblock In \emph{Proceedings of the 57th Annual Meeting of the Association for Computational Linguistics}, 2019.

\bibitem[Zhang et~al.(2023)Zhang, Li, Zhang, Zhan, Wang, Zhou, and Qiu]{zhang2023speechgpt}
Zhang, D., Li, S., Zhang, X., Zhan, J., Wang, P., Zhou, Y., and Qiu, X.
\newblock Speechgpt: Empowering large language models with intrinsic cross-modal conversational abilities, 2023.

\end{thebibliography}
\bibliographystyle{mlsys2025}

\newpage
\appendix
\onecolumn
\section{GPTVQ Algorithm details}\label{app:gptvq_details}

\vspace{-0.2cm}
\paragraph{Codebook update}\label{sec:codebook_update}
After the procedure in Algorithm \ref{gptvq_algo_flow} is complete, we found that the output reconstruction error can be further reduced through a \emph{codebook update}.
Recall that, in line 4 of Algorithm \ref{gptvq_algo_single_group}, $\textbf{Q}$ is incrementally constructed from the elements of $\textbf{C}$. Since this construction constitutes a lookup of values in $\textbf{C}$, the layer-wise objective can still be minimized w.r.t $\textbf{C}$. The objective is a quadratic program and is convex:

\begin{equation}\label{eq:output_mse}
\min_{\textbf{C}_{0},\dots,\textbf{C}_{N}} 
||\mathbf{W}\mathbf{X}-\mathbf{Q}\mathbf{X}||_F^2,
\end{equation}

where $\mathbf{Q}(\textbf{C}_{0},\dots,\textbf{C}_{N})$ is a look-up operation, reconstructing the quantized weights from the centroids. While this objective can be minimized in a closed form, we find that the PyTorch implementation using gradient descent is considerably faster while the solutions are equally good.
The gradient of $\textbf{Q}$ w.r.t. $\textbf{C}$ can be defined simply, as constructing $Q$ only involves a look-up operation.
In each GD step, the values in $\textbf{C}$ are updated, and $\textbf{Q}$ is reconstructed using the new values in $\textbf{C}$, keeping the indices fixed.

\section{Further on-device results}
To investigate the effect of VQ quantized models on model DRAM footprint and latency, we implemented optimized kernels for both a popular mobile CPU architecture, and
Nvidia\textregistered{} GeForce RTX 3080 GPU.

The CPU kernel employs the table lookup ($\texttt{TBL}$) instruction to translate an index of (at most) 5 bits to an 8 bit integer, with two $\texttt{TBL}$ instructions chained for 2D VQ. 
On GPU, we use native CUDA vector types to load and unload data quickly from GPU memory into the registers and back, such as  
\texttt{char4}/\texttt{uchar4}, and custom agglomerations of those, up to \texttt{char128}. 
The code for these kernels will be made available in the future.

We measure the time to transfer and unpack/decode the weights of a Llamav2-7B $\texttt{gate\_proj}$ layer 
($11008 \times 4096$),
for VQ and to uniformly quantized data, and also FP16 on GPU.
Furthermore, we integrate our CPU kernel with a matmul operation for an end-to-end token generation experiment on Llamav2-7B quantized using 1D VQ.

Table~\ref{tab:latency} 
shows that for both data transfer and token generation, 
VQ can achieve significant footprint reductions, with strictly positive latency impact on 
 CPU, and negligible to positive latency impact on Nvidia\textregistered{} GPU.

\begin{table}[ht!]
\centering
\vspace{-.25cm}
\footnotesize
\setlength{\tabcolsep}{5pt}
\caption{\textbf{
Measured VQ data transfer/decoding, and LLM token generation on mobile device.} Exp: experiment, Data Transfer (T) or Token Generation (G). Ptfm: platform, Mobile CPU or NVIDIA\textregistered{} GPU. Format: either Uniform or VQ. Rel. FP: relative footprint. Rel. lat: relative latency.
}\label{tab:latency}
    \vspace{0.2cm}
\begin{tabular}{ccccccc}
    \toprule

    Exp & Ptfm & bpv & Format & $d$ & Rel. FP $\downarrow$ &  Rel. lat. $\downarrow$ 
    \\ 
    \midrule 
    T &  CPU & 4 & Unif & 1D & 1.00$\times$ & 1.00$\times$ \\
    T &  CPU & 8 & Unif & 1D & 2.00$\times$ & 1.93$\times$ \\
    \midrule
    T &  CPU & 3 & VQ & 2D & 0.75$\times$ & 0.98$\times$ \\
    T &  CPU & 2.75 & VQ & 2D & 0.69$\times$ & 0.96$\times$ \\
    T &  CPU & 2.25 & VQ & 2D & 0.56$\times$ & 0.87$\times$\\
    \midrule
    G &  CPU &  3.125 & VQ & 1D & 0.78$\times$ & 0.96$\times$ \\
    \midrule
    \midrule
    T &  GPU & 4 & Unif & 1D & 1.00$\times$ & 1.00$\times$ \\
    T &  GPU & 8 & Unif & 1D & 2.00$\times$ &  1.47$\times$ \\
    T &  GPU & 16 & FP & 1D & 4.00$\times$ & 2.72$\times$ \\ 
    \midrule
    T &  GPU & 2.125 & VQ & 2D & 0.53$\times$ & 1.03$\times$ \\
    T & GPU & 2.125 & VQ & 4D & 0.53$\times$ & 0.71$\times$ \\
    T & GPU & 3.125 & VQ & 2D & 0.78$\times$ & 1.06$\times$ \\
    \bottomrule
\end{tabular}
\vspace{-.5cm}
\end{table}

\section{VQ Configurations}\label{app:configurations}
\begin{table}
\centering
\caption{\textbf{VQ configurations.} Group shape $(r \times c)$ indicates (rows$\times$columns)}\label{tab:configurations}
\begin{tabular}{cccrlc}
    \toprule
    bpv & $d$ & $b$ & group size & group shape & codebook bw\\
    \midrule
    2.125 & 1D & 2 & 256 &  (1$\times$256) & 8\\
    2.125 & 2D & 2 & 2,048 & (4$\times$256) & 8\\
    2.125 & 4D & 2 & 65,536 & (256$\times$256) & 8\\
    \midrule
    2.25 & 1D & 2 & 128 & (1$\times$128) & 8\\
    2.25 & 2D & 2 & 1,024 & (4$\times$256) & 8\\
    2.25 & 4D & 2 & 32,768 & (128$\times$256) & 8\\
    \midrule
    2.75 & 2D & 2.5 & 2,048 & (4$\times$256) & 8\\
    3    & 2D & 2.5 & 512 & (2$\times$256) & 8\\
    \midrule
    3.125 & 1D & 3 & 8,192 & (32$\times$256)  & 8\\
    3.125 & 2D & 3 & 32,768 & (128$\times$256)  & 8\\
    \midrule
    4.125 & 1D & 4 & 1,024 & (4 $\times$256)  & 8\\
    4.125 & 2D & 4 & 65,536 & (256$\times$256) & 8\\
    \bottomrule
\end{tabular}
\end{table}
Table~\ref{tab:configurations} details the studied VQ configurations.

\section{Extended results}\label{app:extra_results}
\begin{table*}[t]
    \setlength{\tabcolsep}{6pt}
    \small
    \centering
    \caption{\textbf{Weight-only quantization results of Llama-1}. We report WikiText2 perplexity in this table; lower is better.}
    \label{tab:llama1_weight_only}
    \renewcommand{\arraystretch}{0.85}
    \begin{tabular}{llccccccccc}
        \toprule
          \multicolumn{2}{l}{
          } & L1-7B & L1-13B & L1-30B & L1-65B \\  \midrule
        FP16 &   & 5.68 & 5.09 & 4.10 & 3.53 \\ 
        \midrule
        \multirow{6}{*}{\shortstack{2.125 bpv\\(W2@g128)}} 
         & RTN &  1.9e3  & 781.20 & 68.04 & 15.08  \\
         & GPTQ &  44.01 & 15.60 & 10.92 & 9.51 \\
         & AWQ &  2.6e5 & 2.8e5 & 2.4e5 & 7.4e4 \\
         & OmniQuant  & 9.72 & 7.93 & 7.12 & 5.95 \\
         
         & \cellcolor{Gray}\textbf{GPTVQ 1D (ours)} 
            & \cellcolor{Gray} 16.29
            & \cellcolor{Gray} \textbf{6.93}
            & \cellcolor{Gray} \textbf{6.04} 
            & \cellcolor{Gray} \textbf{5.19} \\
            
         & \cellcolor{Gray}\textbf{GPTVQ 2D (ours)} 
			& \cellcolor{Gray} \textbf{9.64}
			& \cellcolor{Gray} \textbf{6.58}
			& \cellcolor{Gray} \textbf{5.63}
			& \cellcolor{Gray} \textbf{4.91} \\
        \midrule
        \multirow{6}{*}{\shortstack{2.25 bpv\\(W2@g64)}} & RTN &  188.32 & 101.87 & 19.20 & 9.39  \\
         & GPTQ & 22.10 & 10.06 &  8.54 & 8.31 \\
         & AWQ &  2.5e5 & 2.7e5 & 2.3e5 & 7.4e4 \\
         & OmniQuant  & 8.90  & 7.34 & 6.59 & 5.65 \\ 
         & \cellcolor{Gray}\textbf{GPTVQ 1D (ours)} 
			& \cellcolor{Gray} 16.64
			& \cellcolor{Gray} \textbf{6.78}
			& \cellcolor{Gray} \textbf{5.97}
			& \cellcolor{Gray} \textbf{5.05} \\
			
         & \cellcolor{Gray}\textbf{GPTVQ 2D (ours)} 
			& \cellcolor{Gray} 9.90
			& \cellcolor{Gray} \textbf{6.43}
			& \cellcolor{Gray} \textbf{5.56}
			& \cellcolor{Gray} \textbf{4.86} \\
			
        & \cellcolor{Gray}\textbf{GPTVQ 4D (ours)} 
			& \cellcolor{Gray} \textbf{8.76}
			& \cellcolor{Gray} \textbf{6.33}
			& \cellcolor{Gray} \textbf{5.42}
			& \cellcolor{Gray} \textbf{4.74} \\
			
        \midrule
        \multirow{6}{*}{\shortstack{3.125 bpv\\(W3@g128)}} & RTN & 7.01 & 5.88 & 4.87 & 4.24 \\
         & GPTQ &  6.55 & 5.62 & 4.80 & 4.17 \\
         & AWQ &  6.46 & 5.51 & 4.63 & 3.99 \\
         & OmniQuant  & 6.15 & 5.44 & 4.56 & 3.94 \\ 
          & \cellcolor{Gray}\textbf{GPTVQ 1D (ours)} 
			& \cellcolor{Gray} 6.60
			& \cellcolor{Gray} \textbf{5.34}
			& \cellcolor{Gray} \textbf{4.48}
			& \cellcolor{Gray} \textbf{3.85} \\
			
          & \cellcolor{Gray}\textbf{GPTVQ 2D (ours)} 
			& \cellcolor{Gray} \textbf{6.32}
			& \cellcolor{Gray} \textbf{5.31}
			& \cellcolor{Gray} \textbf{4.38}
			& \cellcolor{Gray} \textbf{3.79} \\
		\bottomrule
    \end{tabular}
    \label{tab:results_weight_only_low_bit}
\end{table*}

\begin{table*}[t]
    \small
    \centering
    \caption{\textbf{Weight-only 4 bit quantization results of Llama-1, Llama-2, and Mistral-7B models}. We report WikiText2 perplexity in this table; lower is better. Models marked `L1' or `L2' denote Llama-v1 and Llama-v2, respectively. M-7B denotes Mistral.}\label{tab:extra_ppl_4bit}
    \renewcommand{\arraystretch}{0.85}
    \setlength\tabcolsep{2pt}
    \begin{tabular}{llcccccccccc}
        \toprule
          \multicolumn{2}{l}{
          } & L1-7B & L1-13B & L1-30B & L1-65B & L2-7B & L2-13B & L2-70B & M-7B & L3-8B\\  \midrule
        FP16 &   & 5.68 & 5.09 & 4.10 & 3.53 & 5.47 & 4.88 & 3.31 & 5.25 & 6.5 \\ 
         \midrule
         
         \multirow{6}{*}{\shortstack{4.125 bpv\\(W4@g128)}} & RTN & 5.96 & 5.25 & 4.23 & 3.67 & 5.72 & 4.98 & 3.46  & 5.42 & 6.73 \\ 
         & GPTQ & 5.85 & 5.20 & 4.23 & 3.65 & 5.61 & 4.98 & 3.42 & 5.35 & 7.32 \\ 
         & AWQ & 5.81 & 5.20 & 4.21 & 3.62 & 5.62 & 4.97 & - & - & - \\ 
         & OmniQuant & \textbf{5.77} & 5.17 & 4.19 & 3.62 & \textbf{5.58} & 6.5 & \textbf{4.95} & - & - \\ 
          & \cellcolor{Gray}\textbf{GPTVQ 1D (ours)} 
			& \cellcolor{Gray} 5.96
			& \cellcolor{Gray} \textbf{5.15}
			& \cellcolor{Gray} \textbf{4.18}
			& \cellcolor{Gray} \textbf{3.60}
			& \cellcolor{Gray} 5.62
			& \cellcolor{Gray} \textbf{4.97}
			& \cellcolor{Gray} \textbf{3.39}
			& \cellcolor{Gray} \textbf{5.32}
            & \cellcolor{Gray} -\\
          & \cellcolor{Gray}\textbf{GPTVQ 2D (ours)} 
			& \cellcolor{Gray} 5.94
			& \cellcolor{Gray} 5.20
			& \cellcolor{Gray} \textbf{4.18}
			& \cellcolor{Gray} 3.64
			& \cellcolor{Gray} \textbf{5.59}
			& \cellcolor{Gray} \textbf{4.94}
			& \cellcolor{Gray} \textbf{3.38}
			& \cellcolor{Gray} \textbf{5.32}
            & \cellcolor{Gray} \textbf{6.41}\\

   		\bottomrule
    \end{tabular}
    \label{tab:results_weight_only_4_bit}
\end{table*}
        
\begin{table*}[t]
    \setlength{\tabcolsep}{3pt}
    \footnotesize
    \centering
    \caption{\textbf{LM-eval results of quantized Llama-v2 7B and 13B, and Llama-v3 8B models.} }\label{tab:lm_eval_results_llama}
    \begin{tabular}{lclccccccc}
        \toprule
          \multicolumn{1}{l}{\textbf{}} & \#Bits & Method  & PIQA  & ARC-e & Arc-c & BoolQ & HellaSwag & Winogrande & \textbf{Avg.}$\uparrow$ \\  
          
    \midrule
    
            \multirow{15}{*}{\shortstack{Llama-v2-7B}} &
            \multicolumn{2}{c}{FP16} &  79.11 &   74.58 &        46.25 &  77.74 &    75.99 &     69.14 &  70.47 \\\cline{2-10}
            
            & \multirow{5}{*}{\shortstack{2.125 bpv\\(W2@g128)}} & RTN          &  51.09 &   27.95 &        25.00 &  41.13 &    26.57 &     49.88 &  36.94 \\
            & & GPTQ         &  54.84 &   30.64 &        25.09 &  53.43 &    33.09 &     51.54 &  41.44 \\
            &  & \cellcolor{Gray} \textbf{VQ-1D} & \cellcolor{Gray} \textbf{61.21} & \cellcolor{Gray} \textbf{38.76} & \cellcolor{Gray} \textbf{24.66} & \cellcolor{Gray} \textbf{62.78} & \cellcolor{Gray} \textbf{45.78} & \cellcolor{Gray} \textbf{53.83} & \cellcolor{Gray} \textbf{47.84} \\
			&  & \cellcolor{Gray} \textbf{VQ-2D} & \cellcolor{Gray} \textbf{71.33} & \cellcolor{Gray} \textbf{57.41} & \cellcolor{Gray} \textbf{32.94} & \cellcolor{Gray} \textbf{65.60} & \cellcolor{Gray} \textbf{59.85} & \cellcolor{Gray} \textbf{64.72} & \cellcolor{Gray} \textbf{58.64} \\
			&  & \cellcolor{Gray} \textbf{VQ-4D} & \cellcolor{Gray} \textbf{73.34} & \cellcolor{Gray} \textbf{60.44} & \cellcolor{Gray} \textbf{34.39} & \cellcolor{Gray} \textbf{65.50} & \cellcolor{Gray} \textbf{63.99} & \cellcolor{Gray} \textbf{65.04} & \cellcolor{Gray} \textbf{60.45} \\\cline{2-10}
            \cline{2-10}

            & \multirow{5}{*}{\shortstack{2.25 bpv\\(W2@g64)}} & RTN          &  58.76 &   36.66 &        24.83 &  41.87 &    40.38 &     51.93 &  42.40 \\
            & & GPTQ         &  60.83 &   39.02 &        25.17 &  59.33 &    45.82 &     55.49 &  47.61 \\
            &  & \cellcolor{Gray} \textbf{VQ-1D} & \cellcolor{Gray} \textbf{64.80} & \cellcolor{Gray} \textbf{49.33} & \cellcolor{Gray} \textbf{28.24} & \cellcolor{Gray} \textbf{65.87} & \cellcolor{Gray} \textbf{53.37} & \cellcolor{Gray} \textbf{54.93} & \cellcolor{Gray} \textbf{52.76} \\
			&  & \cellcolor{Gray} \textbf{VQ-2D} & \cellcolor{Gray} \textbf{72.36} & \cellcolor{Gray} \textbf{63.47} & \cellcolor{Gray} \textbf{35.41} & \cellcolor{Gray} \textbf{72.14} & \cellcolor{Gray} \textbf{60.92} & \cellcolor{Gray} \textbf{64.72} & \cellcolor{Gray} \textbf{61.50} \\
			&  & \cellcolor{Gray} \textbf{VQ-4D} & \cellcolor{Gray} \textbf{73.99} & \cellcolor{Gray} \textbf{64.73} & \cellcolor{Gray} \textbf{36.77} & \cellcolor{Gray} \textbf{71.19} & \cellcolor{Gray} \textbf{64.84} & \cellcolor{Gray} \textbf{65.75} & \cellcolor{Gray} \textbf{62.88} \\\cline{2-10}

            & \multirow{4}{*}{\shortstack{3.125 bpv\\(W3@g128)}} & RTN          &  76.77 &   70.50 &        42.92 &  71.71 &    73.96 &     67.64 &  67.25 \\
            & & GPTQ         &  77.37 &   68.14 &        40.70 &  71.04 &    72.50 &     67.25 &  66.16 \\
            &  & \cellcolor{Gray} \textbf{VQ-1D} & \cellcolor{Gray} \textbf{77.86} & \cellcolor{Gray} \textbf{68.64} & \cellcolor{Gray} \textbf{40.96} & \cellcolor{Gray} \textbf{73.85} & \cellcolor{Gray} \textbf{72.29} & \cellcolor{Gray} \textbf{67.80} & \cellcolor{Gray} \textbf{66.90} \\
			&  & \cellcolor{Gray} \textbf{VQ-2D} & \cellcolor{Gray} \textbf{77.64} & \cellcolor{Gray} \textbf{73.15} & \cellcolor{Gray} \textbf{43.17} & \cellcolor{Gray} \textbf{74.22} & \cellcolor{Gray} \textbf{72.61} & \cellcolor{Gray} \textbf{69.06} & \cellcolor{Gray} \textbf{68.31} \\
        \midrule

        \multirow{15}{*}{\shortstack{Llama-v2-13B}} & \multicolumn{2}{c}{FP16} &  80.52 &   77.53 &        49.23 &  80.52 &    79.38 &     72.14 &  73.22\\\cline{2-10}
            & \multirow{5}{*}{\shortstack{2.125 bpv\\(W2@g128)}}  & RTN          &  58.43 &   32.32 &        25.51 &  47.86 &    39.40 &     48.86 &  42.06 \\
            & & GPTQ         &  59.52 &   40.15 &        27.65 &  57.06 &    41.56 &     53.43 &  46.56 \\
            &  & \cellcolor{Gray} \textbf{VQ-1D} & \cellcolor{Gray} \textbf{73.23} & \cellcolor{Gray} \textbf{64.10} & \cellcolor{Gray} \textbf{35.75} & \cellcolor{Gray} \textbf{71.38} & \cellcolor{Gray} \textbf{60.71} & \cellcolor{Gray} \textbf{65.43} & \cellcolor{Gray} \textbf{61.77} \\
			&  & \cellcolor{Gray} \textbf{VQ-2D} & \cellcolor{Gray} \textbf{75.24} & \cellcolor{Gray} \textbf{68.27} & \cellcolor{Gray} \textbf{38.99} & \cellcolor{Gray} \textbf{69.91} & \cellcolor{Gray} \textbf{65.81} & \cellcolor{Gray} \textbf{68.98} & \cellcolor{Gray} \textbf{64.53} \\
			&  & \cellcolor{Gray} \textbf{VQ-4D} & \cellcolor{Gray} \textbf{75.46} & \cellcolor{Gray} \textbf{71.93} & \cellcolor{Gray} \textbf{42.92} & \cellcolor{Gray} \textbf{67.86} & \cellcolor{Gray} \textbf{69.26} & \cellcolor{Gray} \textbf{66.93} & \cellcolor{Gray} \textbf{65.73} \\
            \cline{2-10}

            &  \multirow{5}{*}{\shortstack{2.25 bpv\\(W2@g64)}} & RTN          &  61.59 &   41.58 &        25.43 &  49.79 &    48.24 &     51.85 &  46.41 \\
            & & GPTQ         &  70.13 &   56.65 &        31.57 &  51.10 &    56.62 &     58.88 &  54.16 \\
            &  & \cellcolor{Gray} \textbf{VQ-1D} & \cellcolor{Gray} \textbf{72.36} & \cellcolor{Gray} \textbf{67.63} & \cellcolor{Gray} \textbf{37.37} & \cellcolor{Gray} \textbf{74.13} & \cellcolor{Gray} \textbf{62.89} & \cellcolor{Gray} \textbf{65.27} & \cellcolor{Gray} \textbf{63.28} \\
			&  & \cellcolor{Gray} \textbf{VQ-2D} & \cellcolor{Gray} \textbf{74.97} & \cellcolor{Gray} \textbf{67.63} & \cellcolor{Gray} \textbf{40.53} & \cellcolor{Gray} \textbf{69.24} & \cellcolor{Gray} \textbf{67.11} & \cellcolor{Gray} \textbf{69.30} & \cellcolor{Gray} \textbf{64.80} \\
			&  & \cellcolor{Gray} \textbf{VQ-4D} & \cellcolor{Gray} \textbf{76.66} & \cellcolor{Gray} \textbf{69.87} & \cellcolor{Gray} \textbf{43.00} & \cellcolor{Gray} \textbf{74.68} & \cellcolor{Gray} \textbf{70.81} & \cellcolor{Gray} \textbf{69.69} & \cellcolor{Gray} \textbf{67.45} \\\cline{2-10}

            &  \multirow{4}{*}{\shortstack{3.125 bpv\\(W3@g128)}} & RTN          &  78.89 &   74.28 &        46.76 &  77.25 &    76.51 &     70.80 &  70.75 \\
            & & GPTQ         &  79.33 &   75.84 &        47.01 &  78.90 &    77.16 &     70.40 &  71.44 \\
            &  & \cellcolor{Gray} \textbf{VQ-1D} & \cellcolor{Gray} \textbf{78.94} & \cellcolor{Gray} \textbf{75.04} & \cellcolor{Gray} \textbf{46.76} & \cellcolor{Gray} \textbf{79.42} & \cellcolor{Gray} \textbf{75.85} & \cellcolor{Gray} \textbf{72.45} & \cellcolor{Gray} \textbf{71.41} \\
			&  & \cellcolor{Gray} \textbf{VQ-2D} & \cellcolor{Gray} \textbf{79.27} & \cellcolor{Gray} \textbf{74.33} & \cellcolor{Gray} \textbf{46.67} & \cellcolor{Gray} \textbf{77.40} & \cellcolor{Gray} \textbf{77.21} & \cellcolor{Gray} \textbf{72.45} & \cellcolor{Gray} \textbf{71.22} \\

        \midrule

        \multirow{15}{*}{\shortstack{Llama-v3-8B}} & \multicolumn{2}{c}{FP16} & 79.9 & 80.1 & 50.4 &  - &    60.2 & 72.8 & 68.6\\\cline{2-10}
            & \multirow{5}{*}{\shortstack{2.125 bpv\\(W2@g128)}}  & RTN          &  53.1 & 24.8 & 22.1 & - &  26.9 & 53.1 & 36.0 \\
            & & GPTQ         &  53.9 & 28.8 & 19.9 &  - &    27.7 & 50.5 & 36.2 \\
            &  & \cellcolor{Gray} \textbf{VQ-1D} & \cellcolor{Gray} \textbf{56.58} & \cellcolor{Gray} \textbf{35.10} & \cellcolor{Gray} \textbf{18.26} & \cellcolor{Gray} \textbf{60.00} & \cellcolor{Gray} \textbf{38.25} & \cellcolor{Gray} \textbf{57.06} & \cellcolor{Gray} \textbf{41.05} \\
			&  & \cellcolor{Gray} \textbf{VQ-2D} & \cellcolor{Gray} \textbf{69.48} & \cellcolor{Gray} \textbf{62.58} & \cellcolor{Gray} \textbf{29.01} & \cellcolor{Gray} \textbf{72.29} & \cellcolor{Gray} \textbf{43.05} & \cellcolor{Gray} \textbf{65.51} & \cellcolor{Gray} \textbf{53.93} \\
			&  & \cellcolor{Gray} \textbf{VQ-4D} & \cellcolor{Gray} \textbf{71.93} & \cellcolor{Gray} \textbf{69.19} & \cellcolor{Gray} \textbf{32.68} & \cellcolor{Gray} \textbf{69.45} & \cellcolor{Gray} \textbf{45.62} & \cellcolor{Gray} \textbf{67.17} & \cellcolor{Gray} \textbf{57.32} \\\cline{2-10}

            &  \multirow{3}{*}{\shortstack{2.25 bpv\\(W2@g64)}} 
            &  \cellcolor{Gray} \textbf{VQ-1D} & \cellcolor{Gray} \textbf{71.16} & \cellcolor{Gray} \textbf{70.24} & \cellcolor{Gray} \textbf{34.04} & \cellcolor{Gray} \textbf{74.13} & \cellcolor{Gray} \textbf{45.71} & \cellcolor{Gray} \textbf{65.27} & \cellcolor{Gray} \textbf{57.29} \\
			&  & \cellcolor{Gray} \textbf{VQ-2D} & \cellcolor{Gray} \textbf{74.27} & \cellcolor{Gray} \textbf{71.30} & \cellcolor{Gray} \textbf{37.54} & \cellcolor{Gray} \textbf{69.24} & \cellcolor{Gray} \textbf{49.07} & \cellcolor{Gray} \textbf{69.30} & \cellcolor{Gray} \textbf{60.30} \\
			&  & \cellcolor{Gray} \textbf{VQ-4D} & \cellcolor{Gray} \textbf{75.68} & \cellcolor{Gray} \textbf{72.60} & \cellcolor{Gray} \textbf{41.04} & \cellcolor{Gray} \textbf{74.68} & \cellcolor{Gray} \textbf{52.22} & \cellcolor{Gray} \textbf{69.69} & \cellcolor{Gray} \textbf{62.25} \\\cline{2-10}

            &  \multirow{4}{*}{\shortstack{3.125 bpv\\(W3@g128)}} & RTN          &  62.3 & 32.1 & 22.5 &  - &    29.1 & 54.7 &40.2 \\
           &  & \cellcolor{Gray} \textbf{VQ-1D} & \cellcolor{Gray} \textbf{77.31} & \cellcolor{Gray} \textbf{77.90} & \cellcolor{Gray} \textbf{43.43} & \cellcolor{Gray} \textbf{79.42} & \cellcolor{Gray} \textbf{57.28} & \cellcolor{Gray} \textbf{72.45} & \cellcolor{Gray} \textbf{65.68} \\
			&  & \cellcolor{Gray} \textbf{VQ-2D} & \cellcolor{Gray} \textbf{77.80} & \cellcolor{Gray} \textbf{76.68} & \cellcolor{Gray} \textbf{45.14} & \cellcolor{Gray} \textbf{77.40} & \cellcolor{Gray} \textbf{58.16} & \cellcolor{Gray} \textbf{72.45} & \cellcolor{Gray} \textbf{66.05} \\

        \bottomrule
    \end{tabular}
    \vspace{-1em}
\end{table*}

\begin{table*}[t]
    \setlength{\tabcolsep}{3pt}
    \footnotesize
    \centering
    \caption{\textbf{LM-eval results of quantized Mistral-7B and Mixtral-8x7B models.} }\label{tab:lm_eval_results_mistral}
    \begin{tabular}{lclccccccc}
        \toprule
          \multicolumn{1}{l}{\textbf{}} & \#Bits & Method  & PIQA  & ARC-e & Arc-c & BoolQ & HellaSwag & Winogrande & \textbf{Avg.}$\uparrow$ \\

        \midrule
            \multirow{15}{*}{\shortstack{Mistral-7B}} & \multicolumn{2}{c}{FP16} &  82.10 &   79.59 &        53.92 &  83.58 &    81.07 &     73.88 &  75.69\\\cline{2-10}
            & \multirow{5}{*}{\shortstack{2.125 bpv\\(W2@g128)}}  & RTN          &  53.05 &   29.42 &        26.62 &  38.56 &    29.26 &     49.57 &  37.75 \\
            & & GPTQ         &  57.73 &   35.65 &        26.62 &  46.06 &    36.06 &     49.49 &  41.93 \\
            &  & \cellcolor{Gray} \textbf{VQ-1D} & \cellcolor{Gray} \textbf{55.22} & \cellcolor{Gray} \textbf{35.94} & \cellcolor{Gray} \textbf{25.51} & \cellcolor{Gray} \textbf{54.01} & \cellcolor{Gray} \textbf{34.35} & \cellcolor{Gray} \textbf{52.01} & \cellcolor{Gray} \textbf{42.84} \\
			&  & \cellcolor{Gray} \textbf{VQ-2D} & \cellcolor{Gray} \textbf{73.78} & \cellcolor{Gray} \textbf{69.02} & \cellcolor{Gray} \textbf{37.80} & \cellcolor{Gray} \textbf{76.57} & \cellcolor{Gray} \textbf{64.52} & \cellcolor{Gray} \textbf{65.35} & \cellcolor{Gray} \textbf{64.51} \\
			&  & \cellcolor{Gray} \textbf{VQ-4D} & \cellcolor{Gray} \textbf{75.90} & \cellcolor{Gray} \textbf{71.63} & \cellcolor{Gray} \textbf{41.98} & \cellcolor{Gray} \textbf{69.85} & \cellcolor{Gray} \textbf{68.59} & \cellcolor{Gray} \textbf{66.46} & \cellcolor{Gray} \textbf{65.73} \\\cline{2-10}

            & \multirow{5}{*}{\shortstack{2.25 bpv\\(W2@g64)}} & RTN          &  60.72 &   38.47 &        27.56 &  44.83 &    46.10 &     51.07 &  44.79 \\
            & & GPTQ         &  65.83 &   46.21 &        30.20 &  62.11 &    50.64 &     55.56 &  51.76 \\
            &  & \cellcolor{Gray} \textbf{VQ-1D} & \cellcolor{Gray} \textbf{67.41} & \cellcolor{Gray} \textbf{59.01} & \cellcolor{Gray} \textbf{33.79} & \cellcolor{Gray} \textbf{67.74} & \cellcolor{Gray} \textbf{53.80} & \cellcolor{Gray} \textbf{55.96} & \cellcolor{Gray} \textbf{56.28} \\
			&  & \cellcolor{Gray} \textbf{VQ-2D} & \cellcolor{Gray} \textbf{74.86} & \cellcolor{Gray} \textbf{69.23} & \cellcolor{Gray} \textbf{40.53} & \cellcolor{Gray} \textbf{74.07} & \cellcolor{Gray} \textbf{65.93} & \cellcolor{Gray} \textbf{67.40} & \cellcolor{Gray} \textbf{65.34} \\
			&  & \cellcolor{Gray} \textbf{VQ-4D} & \cellcolor{Gray} \textbf{76.61} & \cellcolor{Gray} \textbf{73.15} & \cellcolor{Gray} \textbf{42.41} & \cellcolor{Gray} \textbf{77.95} & \cellcolor{Gray} \textbf{69.48} & \cellcolor{Gray} \textbf{69.30} & \cellcolor{Gray} \textbf{68.15} \\\cline{2-10}

            &\multirow{4}{*}{\shortstack{3.125 bpv\\(W3@g128)}} & RTN          &  80.79 &   74.62 &        48.46 &  80.00 &    78.66 &     68.19 &  71.79 \\
            & & GPTQ         &  79.82 &   75.51 &        49.40 &  81.22 &    77.34 &     70.17 &  72.24 \\
            &  & \cellcolor{Gray} \textbf{VQ-1D} & \cellcolor{Gray} \textbf{78.84} & \cellcolor{Gray} \textbf{75.29} & \cellcolor{Gray} \textbf{47.87} & \cellcolor{Gray} \textbf{79.57} & \cellcolor{Gray} \textbf{75.32} & \cellcolor{Gray} \textbf{69.30} & \cellcolor{Gray} \textbf{71.03} \\
			&  & \cellcolor{Gray} \textbf{VQ-2D} & \cellcolor{Gray} \textbf{81.12} & \cellcolor{Gray} \textbf{78.70} & \cellcolor{Gray} \textbf{51.02} & \cellcolor{Gray} \textbf{82.39} & \cellcolor{Gray} \textbf{78.05} & \cellcolor{Gray} \textbf{72.06} & \cellcolor{Gray} \textbf{73.89} \\
        \midrule
        \multirow{15}{*}{\shortstack{Mixtral-8x7B}} & \multicolumn{2}{c}{FP16} &  83.46 &   73.74 &        55.89 &  84.74 &    82.45 &     75.30 &  75.93\\\cline{2-10}
            & \multirow{5}{*}{\shortstack{2.125 bpv\\(W2@g128)}}  & RTN &  51.90 &   27.27 &        25.85 &  47.98 &    27.07 &     49.64 &  38.29 \\
            & & GPTQ    &  59.79 &   35.44 &        27.30 &  52.08 &    41.80 &     50.83 &  44.54 \\
            &  & \cellcolor{Gray} \textbf{VQ-1D} & \cellcolor{Gray} \textbf{68.93} & \cellcolor{Gray} \textbf{50.93} & \cellcolor{Gray} \textbf{33.02} & \cellcolor{Gray} \textbf{62.51} & \cellcolor{Gray} \textbf{52.52} & \cellcolor{Gray} \textbf{61.17} & \cellcolor{Gray} \textbf{54.85} \\
			&  & \cellcolor{Gray} \textbf{VQ-2D} & \cellcolor{Gray} \textbf{76.39} & \cellcolor{Gray} \textbf{57.87} & \cellcolor{Gray} \textbf{38.91} & \cellcolor{Gray} \textbf{74.95} & \cellcolor{Gray} \textbf{67.03} & \cellcolor{Gray} \textbf{71.03} & \cellcolor{Gray} \textbf{64.36} \\
			&  & \cellcolor{Gray} \textbf{VQ-4D} & \cellcolor{Gray} \textbf{78.13} & \cellcolor{Gray} \textbf{65.57} & \cellcolor{Gray} \textbf{46.42} & \cellcolor{Gray} \textbf{78.59} & \cellcolor{Gray} \textbf{72.40} & \cellcolor{Gray} \textbf{71.11} & \cellcolor{Gray} \textbf{68.70} \\\cline{2-10}

            &  \multirow{5}{*}{\shortstack{2.25 bpv\\(W2@g64)}} & RTN           &  62.08 &   38.68 &        28.41 &  54.46 &    44.40 &     53.12 &  46.86 \\
            & & GPTQ         &  66.05 &   42.93 &        28.58 &  50.12 &    49.59 &     55.41 &  48.78 \\
            &  & \cellcolor{Gray} \textbf{VQ-1D} & \cellcolor{Gray} \textbf{69.42} & \cellcolor{Gray} \textbf{50.55} & \cellcolor{Gray} \textbf{36.09} & \cellcolor{Gray} \textbf{64.95} & \cellcolor{Gray} \textbf{59.51} & \cellcolor{Gray} \textbf{63.93} & \cellcolor{Gray} \textbf{57.41} \\
			&  & \cellcolor{Gray} \textbf{VQ-2D} & \cellcolor{Gray} \textbf{77.42} & \cellcolor{Gray} \textbf{62.12} & \cellcolor{Gray} \textbf{42.66} & \cellcolor{Gray} \textbf{72.39} & \cellcolor{Gray} \textbf{70.74} & \cellcolor{Gray} \textbf{68.90} & \cellcolor{Gray} \textbf{65.71} \\
			&  & \cellcolor{Gray} \textbf{VQ-4D} & \cellcolor{Gray} \textbf{79.16} & \cellcolor{Gray} \textbf{67.68} & \cellcolor{Gray} \textbf{48.04} & \cellcolor{Gray} \textbf{76.09} & \cellcolor{Gray} \textbf{73.43} & \cellcolor{Gray} \textbf{71.11} & \cellcolor{Gray} \textbf{69.25} \\\cline{2-10}

            &  \multirow{4}{*}{\shortstack{3.125 bpv\\(W3@g128)}} & RTN          &  81.50 &   68.77 &        50.60 &  80.92 &    79.71 &     72.93 &  72.40 \\
            & & GPTQ         &  80.85 &   69.32 &        52.05 &  81.35 &    78.40 &     74.43 &  72.73 \\
            &  & \cellcolor{Gray} \textbf{VQ-1D} & \cellcolor{Gray} \textbf{80.90} & \cellcolor{Gray} \textbf{71.34} & \cellcolor{Gray} \textbf{52.73} & \cellcolor{Gray} \textbf{84.83} & \cellcolor{Gray} \textbf{77.62} & \cellcolor{Gray} \textbf{73.64} & \cellcolor{Gray} \textbf{73.51} \\
			&  & \cellcolor{Gray} \textbf{VQ-2D} & \cellcolor{Gray} \textbf{82.59} & \cellcolor{Gray} \textbf{72.94} & \cellcolor{Gray} \textbf{54.86} & \cellcolor{Gray} \textbf{84.46} & \cellcolor{Gray} \textbf{80.61} & \cellcolor{Gray} \textbf{74.82} & \cellcolor{Gray} \textbf{75.05} \\

        \bottomrule
    \end{tabular}
    \vspace{-1em}
\end{table*}
Table~\ref{tab:results_weight_only_low_bit} shows GPTVQ results on Llama-1.
Table~\ref{tab:results_weight_only_4_bit} shows GPTVQ results for 4.125 bpv on various models.
Tables~\ref{tab:lm_eval_results_llama}~and~\ref{tab:lm_eval_results_mistral} present detailed LM-eval results.

\section{Mean and standard deviation over multiple runs}
\label{sec:seeds}
\begin{table}[H]
\centering
\caption{\textbf{Mean and standard deviation over 10 random seeds}. Setting used: Llamav2-7B, 2D VQ, 8-bit codebook.}\label{tab:seeds}
\begin{tabular}{ccccc}
    \toprule
   BPV & Mean and Std. Dev.  \\
    \midrule 
   \multirow{1}{*}{3.125} & $5.82 \pm 0.01$ \\ 
   \multirow{1}{*}{4.125} & $5.59 \pm 0.01$ \\ 
    \bottomrule
\end{tabular}
\end{table}

\subsection{Comparison to AQLM}
\label{app:aqlm}
Additive Quantization for Language Models \cite{egiazarian2024aqlm} (AQLM) is a recent method that also uses vector quantization to compress LLMs to very low effective bit widths and achieves impressive bits per value vs accuracy results, as shown in Table~\ref{tab:other_vq}. 
While both GPTVQ and AQLM employ VQ for LLM compression, our methods differ in several significant ways, which affects inference deployment, compression time, and on-disk model size.

AQLM uses larger vector dimension $d$, with $d=8$, scale their codebooks exponentially in $d$, similar to us. E.g., for 2-bit results AQLM uses codebooks with $2^{15}$ or $2^{16}$ 8-dimensional entries, where each entry is stored in FP16.
While the authors have shown that these configurations can be employed on Nvidia\textregistered{} GPUs, codebooks of these sizes would be harder to employ efficiently on a popular mobile platform. This is caused by the fact that many calls to the (5-bit) \texttt{TBL} instruction would be required, leading to significant additional latency during inference time.
For example, decoding a single 16-bit index to an 8-bit FP16 would require $2\times 8 \times 2^{11}$ 5-to-8-bit lookup tables (LUTs), where each lookup requires $2\times 8 \times 11$ instructions to decode.

The full AQLM algorithm requires significant time to compress models. 
Compressing Llamav2-7B requires up to 35 hours on H100, while GPTVQ takes between 30 minutes and 3 hours on a single H100 GPU.
It should however be noted that our method becomes significantly slower for higher quantization dimensionality, mainly due to the EM codebook initialization.

The long runtime of AQLM is caused in part by a block-wise fine-tuning step. 
This step allows the model to correct intra-layer effects of quantization error. 
While GPTVQ has no mechanism to correct intra-layer error effects, its results are competitive with AQLM.
AQLM without the additional block fine-tuning step (i.e., Table~\ref{tab:other_vq}), achieves a perplexity of 7.49 for the WikiText2 test set on Llama-v2-7B, a degradation of 0.38 point compared to 7.11 for GPTVQ under the same conditions.

\section{Ablations on hyperparameter choices}
\label{sec:hparam_ablations}

\begin{table}[t]
\centering
\caption{\textbf{Effect of EM initialization}. Setting used: Llamav2-7B, 2D 3-bit VQ, blocksize 2048.}\label{tab:kmeans_init}
\begin{tabular}{ccccc}
    \toprule
    Lookup method & BPV & Setting & PPL & Time (s) \\
    \midrule 
    \multirow{2}{*}{1D 3B 1024} & \multirow{2}{*}{3.125} 
      & Mahalanobis & 6.05 & 605 \\ 
    & & K++         & 6.16 & 3328 \\
    \midrule 
    \multirow{2}{*}{2D 3B 16384} & \multirow{2}{*}{3.125} 
      & Mahalanobis & 5.65 & 756 \\
    & & K++         & 5.63 & 3168 \\
    \midrule 
    \multirow{2}{*}{1D 4B 2048} & \multirow{2}{*}{4.125} 
      & Mahalanobis & 5.86 & 1272 \\
    & & K++         & 5.88 & 2116 \\
    \midrule 
    \multirow{2}{*}{2D 4B 65536} & \multirow{2}{*}{4.125} 
      & Mahalanobis & 5.59 & 3816 \\
    & & K++         & 5.57 & 6644 \\
    \bottomrule
\end{tabular}
\end{table}
\begin{table}[t]
\centering
\caption{\textbf{Effect of number of EM interations}. Setting used: BLOOM-560m 2D 3-bit VQ with blocksize 4096, perplexity on WikiText2 test set.}\label{tab:kmeans_iters}
\begin{tabular}{cc}
    \toprule
    EM iterations & PPL \\
    \midrule
    10 & 24.49 \\
    30 & 24.18 \\
    50 & 24.12 \\
    75 & 24.11 \\
    100 &  24.09 \\
    \bottomrule
\end{tabular}
\end{table}
\begin{table}[t]
    \centering
    \caption{\textbf{Effect of codebook fine-tuning on final PPL for Llamav2-7B}.}\label{tab:codebook_finetuning}
    \begin{tabular}{cccccc}
        \toprule
        $d$ & $b$ & gs & Update & PPL & Runtime (s)\\
        \midrule
        \multirow{4}{*}{\shortstack{1}} 
            & \multirow{2}{*}{2} 
            & \multirow{2}{*}{512} & N &  43.14 & 625\\
                & &  & Y & 14.02 & 1857 \\\cline{2-5}
            & \multirow{2}{*}{3}
            & \multirow{2}{*}{\shortstack{1024}} & N & 6.05 & 712 \\
                & & & Y & 6.01 & 1916 \\
        \midrule
        \multirow{4}{*}{\shortstack{2}} 
            & \multirow{2}{*}{2} 
            & \multirow{2}{*}{\shortstack{2048}} & N & 8.64 & 723 \\
                & &  & Y & 8.21 & 1335 \\\cline{2-5}
            & \multirow{2}{*}{3}
            & \multirow{2}{*}{\shortstack{8192}} & N & 5.93 & 1585 \\
                & &  & Y & 5.88 & 2195 \\
        \bottomrule
    \end{tabular}
\end{table}
\begin{table}[t]
    \centering
    \caption{\textbf{Effect of scaling block size on perplexity for Llamav2-7B}. $d$: VQ-dimension; $b$: VQ bitwidth per dimension; gs: block size; Codebooks are quantized to 8 bits.}\label{tab:scale_ablation_block_size}
    \begin{tabular}{ccccccccc}
        \toprule
        $d$ & $b$ & gs &\multicolumn{6}{c}{Scaling BS} \\
            &     & & None & 128 & 64 & 32 & 16 & 8\\
        \midrule
        \multirow{2}{*}{\shortstack{1}} & \multirow{1}{*}{2} & 512 & 14.01 & 16.74 & 2744.9 & 480.8 & 15.36 & 13.79
 \\\cline{2-9} 
        & \multirow{1}{*}{3} & 1024 & 6.02 & 5.97 & 6.00 & 5.87 & 5.82 & 5.72
        \\
        \midrule
        \multirow{2}{*}{\shortstack{2}} & \multirow{1}{*}{2} 
        & 2048 & 8.23 & 8.38 & 8.04 & 7.97 & 7.56 & 6.89
 \\ \cline{2-9}
         & \multirow{1}{*}{3} & 8192 & 5.91 & 5.82 & 5.78 & 5.73 & 5.74 & 5.66
 \\
        \bottomrule
    \end{tabular}
\end{table}
\begin{table}[t]
    \centering
    \caption{\textbf{Effect of scaling on perplexity for different models}. Configurations with equal overhead with or without the scaling are considered. $d$: VQ-dimension; $b$: VQ bitwidth per dimension; gs: block size; Codebooks are assumed to be quantized to 8 bit.}\label{tab:scale_ablation_equal_overhead}
    \begin{tabular}{cccccccc}
        \toprule
        $d$ & $b$ & gs & Scale & Llamav2-7B & Llamav2-13B & Mistral-7B & Mixtral-8x7B\\
        \midrule
        \multirow{4}{*}{\shortstack{1}} & \multirow{2}{*}{2} 
        & 256 & N &  14.01 & 7.34 & 15.03 & 8.56\\
         & & 512 & Y & 171.29 & 7.44 & 87.60 & 8.11 \\\cline{2-8}
         & \multirow{2}{*}{3} & 512 & N & 5.98 & 5.21 & 5.76 & 4.60 \\
         & & 1024 & Y & 6.01 & 5.17 & 5.77 & 4.59 \\
        \midrule
        \multirow{4}{*}{\shortstack{2}} & \multirow{2}{*}{2} 
        & 2048 & N & 8.23 & 6.69 & 10.98 & 6.73 \\
         & & 4096 & Y & 8.49 & 6.50 & 10.28 & 6.37 \\\cline{2-8}
         & \multirow{2}{*}{3} & 8192 & N & 5.91 & 5.19 & 8.63 & 4.52 \\
         & & 16384 & Y & 5.56 & 5.11 & 5.53 & 4.30 \\
        \bottomrule
    \end{tabular}
\end{table}

\paragraph{EM initialization}
\label{sec:em_init}
To find seed centroids for EM initialization, we compare k-Means++ \cite{arthur2007k} to a quick and effective initialization method dubbed \emph{Mahalanobis initialization}.
In the latter method, we initialize EM for a matrix of $N$ $d$-dimensional points $\textbf{X}$ by first  sorting all points by Mahalanobis distance \cite{bishop2006pattern} to the mean of the data, then
sampling $k$ points spaced $\lfloor\frac{N}{k-1} \rceil$ apart from the sorted list.
Intuitively, this method ensures that points are sampled at representative distances from the mean.
Table~\ref{tab:kmeans_init} shows perplexity after GPTVQ for different EM initialization seed methods, and find that Mahalanobis initialization performs comparably to k-Means++, at increased speed.

\paragraph{EM iterations}
We explore the effect of the number of EM initialization iterations on the final perplexity of GPTVQ.
Table~\ref{tab:kmeans_iters} shows that even up to 100 iterations, results keep improving slightly, therefore we use 100 iterations as default.

\paragraph{Codebook update}
Table~\ref{tab:codebook_finetuning} includes an ablation on including codebook updates as described in Section~\ref{sec:codebook_update}. 
We find that, in all cases, updating the codebook after running Algorithm~\ref{gptvq_algo_single_group} improves final perplexity, at the expense of moderately increased (though still reasonable) run time.
We thus include codebook update in all training runs.
\paragraph{Method runtime}
GPTVQ can quantize large language models efficiently. 
Exact runtime 
depends on model, quantization setting (groupsize, bitwidth, VQ dimensionality), and several hyperparameters (EM iterations, codebook update iterations). 
As
An indication of realistic run-times on a single H100: Llamav2-7B takes between 30 minutes and 1 hour, while Llamav2-70B takes between 3 and 11 hours.

\section{Codebook compression}\label{app:codebook_compression}
\begin{table}[ht]
    \centering
    \caption{\textbf{Choices in experimental setup leading to comparable bits per value}. $d$: VQ-dimension; $b$: VQ bitwidth per dimension; gs: block size; Q: 8-bit codebook quantization yes/no; SVD: codebook SVD yes/no. BPV: bits per value.}\label{tab:overhead_choices}
    \begin{tabular}{ccccccc}
        \toprule
        $d$ & $b$ & gs & Q & SVD & BPV & PPL\\
        \midrule
        \multirow{6}{*}{\shortstack{1}} & \multirow{3}{*}{2} & 512 & N & N & 2.125 &  14.01\\
         & & 256 & Y & N & 2.125 & 11.57\\
         & & 256 & N & Y & 2.125 & 44.99 
         \\\cline{2-7}
         & \multirow{3}{*}{3} & 1024 & N & N & 3.125 & 6.01 \\
         & & 512 & Y & N & 3.125 & 5.98 \\
         & & 512 & N & Y & 3.125 & 5.98 \\
        \midrule
        \multirow{4}{*}{\shortstack{2}} & \multirow{2}{*}{2} & 4096 & N & N & 2.125 & 8.37 \\
         & & 2048 & Y & N & 2.125 & 8.23 \\\cline{2-7}
         & \multirow{2}{*}{3} & 16384 & N & N & 3.125 & 5.93 \\
         & & 8192 & Y & N & 3.125 & 5.87\\
        \bottomrule
    \end{tabular}
\end{table}
While we find that 8 bit quantization of codebooks provides best results for the same overhead, we explore a different approach to codebook compression in this section.

For the case where $d=1$, we could further compress the codebook $\textbf{C}$ by stacking all codebooks for multiple blocks (e.g. all blocks in a tensor) and rank-reducing the resulting matrix.
For a single tensor, $\textbf{C}$ has shape $N_G\times k$, where $N_G$ is the number of groups in the corresponding weight tensor, $k$ is the number of centroids per codebook.
We first sort values in each codebook in $\textbf{C}$, and reassign the indices in $\textbf{I}$ accordingly.
Then, we perform SVD on $\textbf{C}$, leading to matrices $\textbf{U}$, $\mathbf{\Sigma}$ and $\textbf{V}$, of shapes $N_G\times k$, $k \times k$ and $k \times k$, respectively.
$\textbf{U}'=\textbf{U}\mathbf{\Sigma}$, and reduce the rank of this matrix to $r$, yielding a $N_G \times r$ shaped matrix $\textbf{U}''$.
We also reduce the rank of $\textbf{V}$ accordingly, yielding $r\times r$ matrix $\textbf{V}'$.
Then, we perform gradient descent (GD) on the loss of equation \ref{eq:output_mse}, but with respect to the codebook tensor factors $\textbf{U}''$ and $\textbf{V}'$. 
In each GD step, $\widehat{\textbf{C}}$ is created as $\widehat{\textbf{C}}=\textbf{U}''\textbf{V}'^T$, and the rest of the codebook up procedure as described earlier is followed.
Lastly, only the codebook tensor factor $\textbf{U}''$ is quantized, as $\textbf{V}'$ gives very little overhead. 
During inference, $\widehat{\textbf{C}}$ is quantized per codebook after construction.

For higher dimensions, Tucker factorization could be employed. However, in this case there is no natural ordering in which to sort the elements of each codebook.

In Table~\ref{tab:overhead_choices} we compare the effect of either rank reducing by 50\%, or quantizing the codebook to 8-bit (our default approach), to keeping the codebook in FP16 and increasing the group size. 
In all three settings the overhead of the codebook is the same.
We see that, for the same overhead, quantization gives best results. For this reason, and because codebook SVD does not easily apply to $d>1$, we have not explored codebook SVD further, and instead use INT8 quantization as our default approach.

\section{Blockwise data normalization}
\label{sec:blockwise_data_normalization}
In order to lower the error of vector quantization, we apply blockwise data normalization to the data before the codebook initialization. For each group corresponding to a new codebook we perform element-wise division $\mathbf{W}_i \oslash \mathbf{S}_i $ of the weight sub-matrix matrix $\mathbf{W}_i$ by the corresponding scales $\mathbf{S}_i$. The scale is computed block-wise for every sub-row of $\mathbf{W}_i$, e.g. for a block size of 16, 32, or 64. 

Given a set of blocks (sub-rows) $\mathbf{w}^{(i)}$, the scale $s^{(i)}$ for each of them is computed as $s^{(i)}=\max_j |w_j^{(i)}|$. In order to minimize the overhead, the scales are quantized to 4-bit integer.

We found that it is beneficial to perform quantization in log-scale to capture several orders of magnitudes in weights. The quantized scales are computed as $s^{(i)}_{int} = \lceil \frac{\log_2[s^{(i)}]-z}{a}\rfloor a$, where $a$ is the quantization scale shared among the group of weights. In order to accurately represent zero in log-space which corresponds to unit scaling, we use the floating point offset $z$. In practice the value of $z$ is shared within the columns of $\mathbf{W}$ and thus has negligible overhead. Finally the scaled sub-row is normalized as $\mathbf{w} \cdot 2^{-a s_{int}-s_0}$, where $s_0=\log_2(z)$. The scaled data is used for codebook initialization. The inverse scaling is applied at VQ decoding step.

\section{Centroid fine-tuning}
\begin{table*}[t]
    \setlength{\tabcolsep}{5pt}
    \small
    \centering
    \caption{\textbf{Weight-only quantization results of Llama-v2, Llama-v3, Mistral, and Phi-3 mini 4k instruct models with centroid fine-tuning (+FT)}. We report WikiText2 perplexity at the context length of 2048 and average zero-shot accuracy. Models marked L2 denote Llama-v2, L3 denote Llama-v3, M denotes Mistral, and P3-mini denotes Phi-3 mini 4k instruct.}
    \vspace{0.2cm}
    \renewcommand{\arraystretch}{0.98}
    \setlength\tabcolsep{3pt}
    \begin{tabular}{llccccc|ccccc}
        \toprule
        & & \multicolumn{5}{c}{WikiText2 perplexity $\downarrow$} &  \multicolumn{5}{c}{Zeroshot avg acc. $\uparrow$}\\
          & & L2-7B & L2-13B & L3-8B & M-7B & P3-mini & L2-7B & L2-13B & L3-8B & M-7B & P3-mini \\  \midrule
        FP16 &   & 5.47 & 4.89 & 6.14 & 5.25 & 6.36 & 70.4 & 73.3 & 74.2 & 75.7 & 76.1 \\ 
        \midrule
        \multirow{4}{*}{\shortstack{2.125\\ \\W2\\g128}}        
        
         & \textbf{GPTVQ 2D} 
			& 7.60
            & 6.38
            & 10.99
            & 7.69
            & 10.18
            & 59.9
            & 65.0
            & 60.3
            & 64.0
            & 62.4
            \\
            
         & \textbf{GPTVQ 2D + FT} 
			& \textbf{6.57}
            & \textbf{5.70}
            & \textbf{8.65}
            & \textbf{6.22}
            & \textbf{8.00}
            & \textbf{64.2}
            & \textbf{67.2}
            & \textbf{66.8}
            & \textbf{69.1}
            & \textbf{67.5}           
            \\
            
        & \textbf{GPTVQ 4D} 
			& 7.14
            & 5.97
            & 9.61
            & 6.75
            & 9.01
            & 62.2
            & 67.4
            & 61.4
            & 67.7
            & 65.3
            \\
            
         & \textbf{GPTVQ 4D + FT} 
			& \textbf{6.39}
            & \textbf{5.58}
            & \textbf{8.25}
            & \textbf{6.04}
            & \textbf{7.75}
            & \textbf{64.4}
            & \textbf{68.4}
            & \textbf{68.9}
            & \textbf{70.0}
            & \textbf{68.4}           
            \\
        \midrule
        \multirow{4}{*}{\shortstack{2.25\\ \\W2\\g64}}        
        
         & \textbf{GPTVQ 2D}
			& 7.42
            & 6.25
            & 10.45
            & 7.36
            & 9.70
            & 60.6
            & 66.7
            & 61.9
            & 64.6
            & 63.9
            \\
            
         & \textbf{GPTVQ 2D + FT} 
			& \textbf{6.49}
            & \textbf{5.65}
            & \textbf{8.51}
            & \textbf{6.16}
            & \textbf{7.91}
            & \textbf{64.1}
            & \textbf{68.7}
            & \textbf{67.1}
            & \textbf{70.4}
            & \textbf{67.8}           
            \\
            
        & \textbf{GPTVQ 4D} 
			& 6.92
            & 5.88
            & 9.26
            & 6.59
            & 8.86
            & 62.0
            & 67.4
            & 63.1
            & 68.2
            & 64.6
            \\
            
         & \textbf{GPTVQ 4D + FT} 
			& \textbf{6.31}
            & \textbf{5.54}
            & \textbf{8.13}
            & \textbf{5.99}
            & \textbf{7.59}
            & \textbf{65.6}
            & \textbf{70.0}
            & \textbf{68.9}
            & \textbf{70.2}
            & \textbf{68.9}           
            \\

        \midrule
        \multirow{4}{*}{\shortstack{3.125\\ \\W2\\g128}}        

        & & & & & & & & & & & \\
         
         & \textbf{GPTVQ 2D} 
			& 5.79
            & 5.11
            & 6.97
            & 5.53
            & 6.93
            & 67.8
            & \textbf{71.4}
            & 71.1
            & \textbf{73.9}
            & 72.3
            \\
            
         & \textbf{GPTVQ 2D + FT} 
			& \textbf{5.73}
            & \textbf{5.10}
            & \textbf{6.96}
            & \textbf{5.49}
            & \textbf{6.56}
            
            & \textbf{68.7}
            & 71.2
            & \textbf{72.5}
            & 73.8
            & \textbf{73.7}           
            \\
        & & & & & & & & & & & \\
        
        \bottomrule
    \end{tabular}
    \vspace{0.1cm}
    \label{tab:vqat_results}
\end{table*}
The GPTVQ algorithm has two reasons for being not optimal.
First, it processes layers sequentially, minimizing the local error in each layer.
Second, the codebook is selected per group that usually spans across multiple columns. Once the first $d$ columns are quantized (step~\ref{lst:line:quant_op} in Algorithm~\ref{gptvq_algo_single_group}), the remaining weights of the \textit{same} group are updated (step~\ref{lst:line:ref_accum_update} in Algorithm~\ref{gptvq_algo_single_group}).
Therefore, the selected codebook will not optimally represent the remaining weights in the group.

One straight-forward solution to improve the performance is to fine-tune the selected codebook entries end-to-end.
To this end, we fine-tuned the codebooks and scales on SlimPajama~\cite{cerebras2023slimpajama} dataset for $1000$ steps with batch size $32$ and sequence lengths of $1024$.
We used Adam optimizer with warmup and cosine scheduler for both codebooks and scales.
We selected the codebook learning rate using a grid search.
The scales learning rate was reduced by $10^3$ times.

Table~\ref{tab:vqat_results} compares the fine-tuning results with the original GPTVQ results.
We can see that for all models and bits, fine-tuning consistently improves the results except zero-shot accuracy at $3.125$ bits.

Note that we also tried to update the indices similarly to PV-Tuning~\cite{malinovskii2024pv}, but we did not see any significant improvement of results.
We hypothesize that due to the nature of GPTVQ update of the weights, the selected indices are very close to the optimal one.
This is partially confirmed by the fact that AQLM algorithm, which is used by PV-tuning, requires layer-wise and block-wise fine-tuning of centroids, while the original GPTVQ algorithm produces comparable results right after initialization.

\end{document}